\documentclass[journal]{IEEEtran}
\usepackage{amsmath,amsfonts}
\usepackage{algorithmic}
\usepackage{algorithm}
\usepackage{array}
\usepackage{subfigure}
\usepackage{textcomp}
\usepackage{stfloats}
\usepackage{url}
\usepackage{verbatim}
\usepackage{graphicx}
\usepackage{cite}
\usepackage{xcolor}         
\usepackage{tcolorbox}

\usepackage{microtype}
\usepackage{graphicx}

\usepackage{booktabs} 
\usepackage{colortbl}
\usepackage{tikz}
\usepackage{pgfplots}
\usepackage{pgfplotstable}
\usepackage{amsmath}
\usepackage{amssymb}
\usepackage{amsthm}
\usepackage{accents}
\usepackage{algorithmic}
\usepackage{algorithm}
\usepackage{array}
\usepackage{cite}

\usepackage{eqparbox}
\usepackage{bm}  
\usepackage[font=small,labelfont=bf]{caption}
\usepackage{multirow}
\usepackage{soul}
\usepackage{hyperref}

\makeatletter
\newcommand{\thickhline}{%
    \noalign {\ifnum 0=`}\fi \hrule height 1pt
    \futurelet \reserved@a \@xhline
}
\newcolumntype{"}{@{\hskip\tabcolsep\vrule width 1pt\hskip\tabcolsep}}
\makeatother

\hypersetup{
    colorlinks,
    linkcolor={red!50!black},
    citecolor={blue!50!black},
    urlcolor={blue!80!black}
}

\usepackage{amsmath}
\usepackage{amssymb}
\usepackage{mathtools}
\usepackage{amsthm}

\usepackage[capitalize,noabbrev]{cleveref}

\theoremstyle{plain}
\newtheorem{theorem}{Theorem}

\newtheorem{lemma}{Lemma}

\theoremstyle{definition}

\newtheorem{assumption}{Assumption}
\theoremstyle{remark}

\usepackage[textsize=tiny]{todonotes}

\usepackage{float}
\usepackage{orcidlink}

\pgfplotstableread{ 
Label     Class0 Class2 Class4 Class6 Class8
Client0   50  50 0 0 0
Client2   0  50 50 0 0
Client4   0  0 50 50 0
Client6  0  0 0 50 50
Client8  50  0 0 0 50
}\supdata
\pgfplotstableread{ 
Label     Class1 Class3 Class5 Class7 Class9
Client1   50  50 0 0 0
Client3   0  50 50 0 0
Client5   0  0 50 50 0
Client7  0  0 0 50 50
Client9  50  0 0 0 50
}\infdata

\begin{document}

\title{FedShift: Robust Federated Learning Aggregation Scheme in Resource Constrained Environment via Weight Shifting}


\author{Jungwon Seo\orcidlink{0000-0001-8174-9352}, Minhoe Kim\orcidlink{0000-0002-3765-1208} and Chunming Rong\orcidlink{0000-0002-8347-0539} \thanks{This work is funded by NCS2030, RCN\#331644, a national research center financed by the Research council of Norway, the industry sponsors, and the academic partners. \textit{(Corresponding author: Minhoe Kim.)}}
\IEEEcompsocitemizethanks{\IEEEcompsocthanksitem J. Seo and C. Rong are with the Department of Electrical Engineering and Computer Science, University of Stavanger, Stavanger 4021, Norway (e-mail: jungwon.seo@uis.no and chunming.rong@uis.no). \IEEEcompsocthanksitem M. Kim is with the Department of Computer Convergence Software, Korea University, Sejong-si 30019, Republic of Korea (e-mail: kimminhoe@korea.ac.kr).}
}

\markboth{IEEE INTERNET OF THINGS JOURNAL,~Vol.~XX, No.~X, XX~20XX}%
{Shell \MakeLowercase{\textit{et al.}}: A Sample Article Using IEEEtran.cls for IEEE Journals}

\IEEEpubid{0000--0000/00\$00.00~\copyright~2021 IEEE}

\maketitle

\begin{abstract}
Federated Learning (FL) commonly relies on a central server to coordinate training across distributed clients. While effective, this paradigm suffers from significant communication overhead, impacting overall training efficiency. To mitigate this, prior work has explored compression techniques such as quantization. However, in heterogeneous FL settings, clients may employ different quantization levels based on their hardware or network constraints, necessitating a mixed-precision aggregation process at the server. This introduces additional challenges, exacerbating client drift and leading to performance degradation. In this work, we propose \texttt{FedShift}, a novel aggregation methodology designed to mitigate performance degradation in FL scenarios with mixed quantization levels. \texttt{FedShift} employs a statistical matching mechanism based on weight shifting to align mixed-precision models, thereby reducing model divergence and addressing quantization-induced bias. Our approach functions as an add-on to existing FL optimization algorithms, enhancing their robustness and improving convergence. Empirical results demonstrate that \texttt{FedShift} effectively mitigates the negative impact of mixed-precision aggregation, yielding superior performance across various FL benchmarks.
\end{abstract}

\begin{IEEEkeywords}
Federated Learning, Quantization, Optimization
\end{IEEEkeywords}

\section{Introduction}
\IEEEPARstart{F}{ederated} Learning (FL) is a machine learning technique designed to enable training on distributed data while ensuring that data remains localized, thereby safeguarding privacy \cite{konecny2016federated, mcmahan2017communication, smith2017federated}. In the standard FL workflow, each client locally trains a model, which often involves neural networks, and sends the resulting model's weights to a central server. The server aggregates these weights to form a global model, which is subsequently redistributed to the clients for further training. Thanks to this approach, FL has gained significant popularity in edge computing and IoT computing, where computing and storage resources are limited~\cite{9475501,9691274,9984824}. Additionally, it is highly beneficial in scenarios where privacy-sensitive information cannot leave the device, ensuring data security and compliance with regulatory requirements~\cite{10375241,9766218,wang2022privacy}.

Despite its advantages, FL suffers from high communication overhead between the server and clients, which becomes more pronounced with the increasing size and complexity of modern neural network-based models~\cite{villalobos2022machine}. This overhead often acts as a bottleneck to training efficiency~\cite{luping2019cmfl, shahid2021communication}, driving the development of methods to reduce the volume of model exchanged. One prominent approach is communication compression via parameter quantization~\cite{reisizadeh2020fedpaq, tonellotto2021neural, tang2018communication, chen2021dynamic, huang2024stochastic, basu2019qsparse, malekijoo2021fedzip, haddadpour2021federated}. Model parameters are encoded to quantized values which is typically compressed from 32-bit floating point to lower-bit format. This reduces model size proportionally while aiming to preserve essential information. The technique is particularly effective in mitigating communication constraints on uplink channels~\cite{konecny2016federated, honig2022dadaquant} in FL, where local models are sent from clients to a central server. 


In practice, FL clients may use different quantization levels to accommodate varying computational capabilities and network conditions~\cite{jhunjhunwala2021adaptive,honig2022dadaquant,mao2022communication}. For instance, clients with limited computing resources or a poor communication link might opt for a low-bit quantization, whereas clients with ample resources or high bandwidth could use high-bit quantization. Although this adaptive quantization level allows more clients to participate in FL by reducing straggler effects\cite{li2020fedprox}, it also leads to the aggregation of mixed-precision models~\cite{yoon2022bitwidth,yuan2024mixed,chen2024mixed}. In every aggregation round, the interplay of these mixed-precision models induces quantization discrepancies~\cite{micikevicius2018mixed}, which may degrade performance.

Mitigating this quantization discrepancy is critical, as FL already contends with a persistent challenge known as client drift. Client drift, stemming from inherent data heterogeneity, has been a fundamental issue in FL and a focal point of extensive research~\cite{zhao2018federated,karimireddy2020scaffold,li2022federated,seo2024understanding}. Quantization discrepancies further exacerbate the unpredictability of client drift, making its effects even more pronounced~\cite{li2023analysis}. While some studies~\cite{haddadpour2021federated,li2023analysis,huang2024stochastic} explore the interplay between quantization and client drift, they predominantly assume uniform quantization levels, leaving mixed-precision aggregation largely unexamined. Furthermore, existing approaches often require additional downlink communication or extra client-side storage to correct for quantization-induced drift, ultimately undermining the communication efficiency gains that quantization aims to provide.

\IEEEpubidadjcol

In this paper, we aim to focus on addressing the aggregation of mixed precision models of FL systems with statistical solution by introducing \texttt{FedShift}, a novel aggregation technique inspired by weight standardization~\cite{qiao2019micro} and batch normalization~\cite{ioffe2015batch}. \texttt{FedShift} leverages the non-quantized parameters to compute statistical adjustments (shifts) that align the distributions of quantized models. By shifting the mean of each parameter distribution, \texttt{FedShift} reduces divergence across mixed-precision updates, stabilizing training in the presence of varying quantization levels. This approach is efficient, requires minimal computation, and is easy to integrate into existing FL pipelines.

In summary, we present the following major contributions:

\begin{itemize}

\item We introduce a novel FL method, \texttt{FedShift}, designed to mitigate the negative effects of quantization and client drift on model performance through weight shifting during the server aggregation stage (see Section~\ref{proposed-fedshift}).
\item We provide a comprehensive theoretical investigation, encompassing both convergence and divergence analyses. The convergence analysis addresses federated algorithms incorporating weight adjustments on the server side, while the divergence analysis illustrates how \texttt{FedShift} achieves more stable global training (see Section~\ref{theoritical-analysis}).
\item Through extensive experiments under various FL and quantization settings, we demonstrate that \texttt{FedShift} consistently enhances performance when integrated with other FL algorithms, especially when communication compression is applied (see Section~\ref{evaluation}).
\end{itemize}

\section{Related Works}\label{related-work}

\noindent\textbf{Client Drift.} Since the introduction of \texttt{FedAvg}~\cite{mcmahan2017communication}, FL research has primarily focused on addressing client drift caused by data heterogeneity. Early representative approaches include \texttt{FedProx}~\cite{li2020fedprox}, which introduced a proximal term in the local loss function to constrain the divergence between global and local models, and \texttt{SCAFFOLD}~\cite{karimireddy2020scaffold}, which corrected client gradient updates to enhance alignment across clients. Beyond client-side interventions, server-side strategies modifying the global model update process have also emerged~\cite{reddi2021adaptive,jhunjhunwala2022fedvarp}. Subsequent works have expanded on these ideas, leveraging loss function modifications and gradient adjustments to develop algorithms that consistently outperform \texttt{FedAvg} under strong non-\textit{i.i.d.} (Independent and Identically Distributed) data heterogeneity.

\noindent\textbf{Communication Compressed FL} While addressing client drift has been a primary focus in improving model performance in FL, research on enhancing the training speed has also been actively pursued. Unlike centralized learning, FL involves communication between clients and the server over potentially heterogeneous network environments, leading to challenges in training efficiency due to system heterogeneity. Among various methods, communication compression techniques have gained significant attention. One foundational approach is \texttt{QSGD}~\cite{alistarh2017qsgd}, initially developed for centralized learning to reduce the gradient size in multi-GPU environments. \texttt{FedPAQ}~\cite{reisizadeh2020fedpaq} extended this technique to FL, introducing gradient quantization to reduce communication overhead. However, to address persistent challenges from client drift caused by data heterogeneity, subsequent works such as \texttt{FedCOMGATE}~\cite{haddadpour2021federated} and \texttt{SCALLION}~\cite{huang2024stochastic} incorporated strategies similar to \texttt{SCAFFOLD}, combining gradient quantization with client-drift mitigation to enhance robustness and efficiency.

\noindent\textbf{Compression Error} 
While quantization enhances communication efficiency, it inherently introduces a loss of information, potentially impacting the performance of the resulting global model. Previous studies primarily addressed this issue by incorporating it into the broader framework of mitigating client drift caused by data heterogeneity. In contrast, \texttt{Fed-EF}~\cite{li2023analysis} explicitly tackled quantization-induced errors using an error feedback mechanism~\cite{stich2018sparsified}. Specifically, during client-side quantization, the residual between pre- and post-quantization values from the previous round is incorporated into the current quantization process to reduce cumulative error.

\noindent\textbf{Mixed Precision.} 
Most studies have applied a uniform quantization level across all clients. However, to account for system heterogeneity and avoid unnecessary quantization, an algorithm has been proposed that dynamically assigns quantization levels tailored to each client~\cite{honig2022dadaquant}. Furthermore, while not explicitly focused on communication efficiency, another study has addressed scenarios where participating devices in FL have varying allowable bit width, introducing strategies to adapt to these constraints~\cite{yoon2022bitwidth} by using trainable weight dequantizer.

\section{Problem Setup}

\subsection{Federated Learning}
The general FL optimization process can be formulated as follows:

\begin{equation}
\min_{\mathbf{w}} F(\mathbf{w}) \triangleq \sum_{k=1}^N p_k F_k(\mathbf{w}),
\end{equation}

where \(\mathbf{w}\) is the parameter vector to be optimized, \(F(\mathbf{w})\) is the global objective function, and \(N\) is the total number of federated clients. \(p_k\) represents the contribution of client \(k\).

The local objective function for client \(k\) is defined as:

\begin{equation}
F_k(\mathbf{w}) \triangleq \frac{1}{n_k} \sum_{j=1}^{n_k} \ell(\mathbf{w}; \xi_{k,j}),
\end{equation}

where \(n_k\) is the number of data samples for client \(k\), and \(\ell(\mathbf{w}; \xi_{k,j})\) is the loss function for the model parameter \(\mathbf{w}\) and the \(j\)-th data sample \(\xi_{k,j}\) of client \(k\).

During each communication round, the local update for client 
\(k\) is conducted using stochastic gradient descent (SGD), initialized with the global model weights $\mathbf{w}^{t}$. The model weights for client \(k\) at local epoch \(e+1\) are updated as:

\begin{equation}
\mathbf{w}_{k}^{t, e+1} = \mathbf{w}_{k}^{t,e} - \eta_l \nabla F_k(\mathbf{w}_{k}^{t,e}, \xi_{k}),
\end{equation}

where \(\eta_l\) is the local learning rate, \(\mathbf{w}_{k}^{t,e}\) represents the weights of client \(k\) at local epoch \(e\), and \(\nabla F_k(\mathbf{w}_{k}^{t,e}, \xi_k)\) is the gradient of the local objective function computed using a mini-batch of data \(\xi_k\). After \(E\) local epochs, the final model weights for client \(k\) are sent back to the server as \(\mathbf{w}_k^{t+1} = \mathbf{w}_k^{t,E}\). Then the server aggregates the clients' models to make a global model which is sent to the clients for the next round. Then, the generic aggregation strategy of participating clients can be represented as $\mathbf{w}^{t+1} = \mathbb{E}_{k|t+1}[\mathbf{w}_k^{t+1}]$, e.g., $\mathbb{E}_{k|t}[\mathbf{w}_k^t]=\sum_k p_k\mathbf{w}_k^t$ for \texttt{FedAvg}.
 The training is completed after \(T\) rounds of aggregations. In the case of partial client participation, the participation ratio \(C\) can be set, resulting in \(C \cdot N\) participating clients per round.

\subsection{System Model}

\begin{figure}[htbp]

\centering
\includegraphics[width=0.4\textwidth]{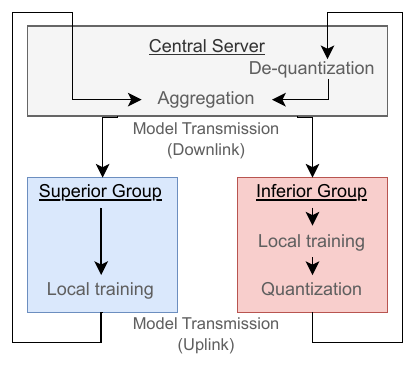}
\caption{Training flow with inferior group incorporating quantization process (mixed-precision)}
\label{fig_archtecture}

\end{figure}

Our system model is demonstrated in Figure~\ref{fig_archtecture}. In this model, we assume that clients with subpar network conditions have been identified and categorized into two groups as follows: \textit{inferior} and \textit{superior}, based on the quality of their network environment. Additionally, our system model incorporates communication compression through weight quantization techniques~\cite{li2022federatedwq,chen2022energy,gimenez2024effects}. Nevertheless, other forms of compression~\cite{bernstein2018signsgd,aji2017sparse} can also be employed.

In this training flow, clients of the \textit{superior} group follow the standard FL flow (e.g., \texttt{FedAvg}). As their network speed is sufficient, it is unnecessary to quantize their model. Conversely, quantization is applied to the \textit{inferior} group. Upon receiving the model from the central server, the \textit{inferior} group starts training and quantizes the trained weights. These quantized weights are then transmitted to the central server. Upon aggregation, model parameters received from the \textit{superior} group are directly utilized without further processing, whereas those from the \textit{inferior} group undergo a dequantization process before integration.  Thus, the aggregation process can be expressed as follows:
\begin{equation}
\mathbf{w}^{t+1}=\frac{1}{|\mathbb{K}|}\left( \sum_{i\in \mathbb{I}}{ DeQ(Q(\mathbf{w}_{i}^{t+1}))}+\sum_{j \in \mathbb{S}} \mathbf{w}_{j}^{t+1} \right)
\end{equation}

where $Q$ and $DeQ$ denote the quantization and dequantization operations, respectively. Here,  $\mathbb{K}$ represents the set of selected clients, while $\mathbb{I}$ and $\mathbb{S}$ correspond to the sets of inferior and superior clients selected at round $t$.

\subsection{Weight Quantization}\label{adx.quantization}

\begin{figure}[!htbp]
\centering
\subfigure[8-bit]{
    \includegraphics[width=0.22\textwidth]{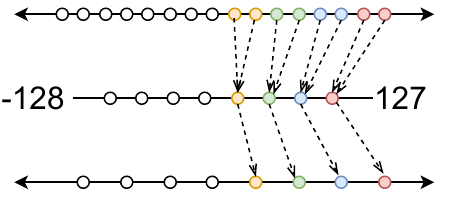}
}
\subfigure[4-bit]{
   \includegraphics[width=0.22\textwidth]{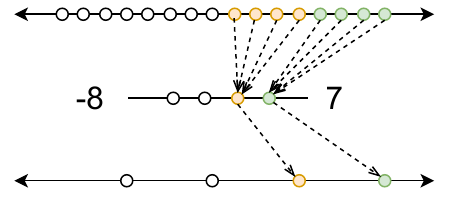}
}
\caption{Illustration of quantization and dequantization process with various bit widths}

\label{fig:quant-dequant}
\end{figure}
In FL, weight quantization is employed to reduce communication overhead by transmitting model weights from clients to the server in a lower-bit representation. During this process, the weights are quantized into fewer bits for transmission and subsequently decompressed or dequantized back into a 32-bit representation at the server for further use~\cite{gimenez2024effects}. As illustrated in Figure~\ref{fig:quant-dequant}, there exists a trade-off between the target quantization bit-width and the resulting information loss. Specifically, lower quantization bit-widths lead to increased information loss upon dequantization, which may impact model performance.

In this study, we employ two quantization techniques to achieve a balance between the computational overhead of quantization and the need for preserving information. First, Uniform quantization is a method of remapping values from the original intervals into evenly spaced and smaller intervals. One of the well-known uniform quantization methods is called range-based asymmetric quantization (ASYM) \cite{mills2019communication,guan2019post}, as shown in Equation \ref{uniform-quantization}, where $\mathbf{w}$ is a parameter of the model and $\mathcal{N}$ is the number of the target bits. The process for dequantization can be derived simply by employing the inverse of the formula as demonstrated in Equation \ref{uniform-dequantization}.

\begin{equation}
\label{uniform-quantization}
Q(\mathbf{w}) = round(\frac{\mathbf{w}-\mathbf{w}_{min}}{\mathbf{w}_{max}-\mathbf{w}_{min}}*(2^\mathcal{N}-1))
\end{equation}
\begin{equation}
\label{uniform-dequantization}
DeQ(\mathbf{w}) = Q(\mathbf{w})*(\frac{\mathbf{w}_{max}-\mathbf{w}_{min}}{2^\mathcal{N}-1})+\mathbf{w}_{min}
\end{equation}

\begin{figure}[!htbp]
\centering\hspace{-8mm} 
\subfigure[Epoch: 0]{    \includegraphics[width=0.28\textwidth]{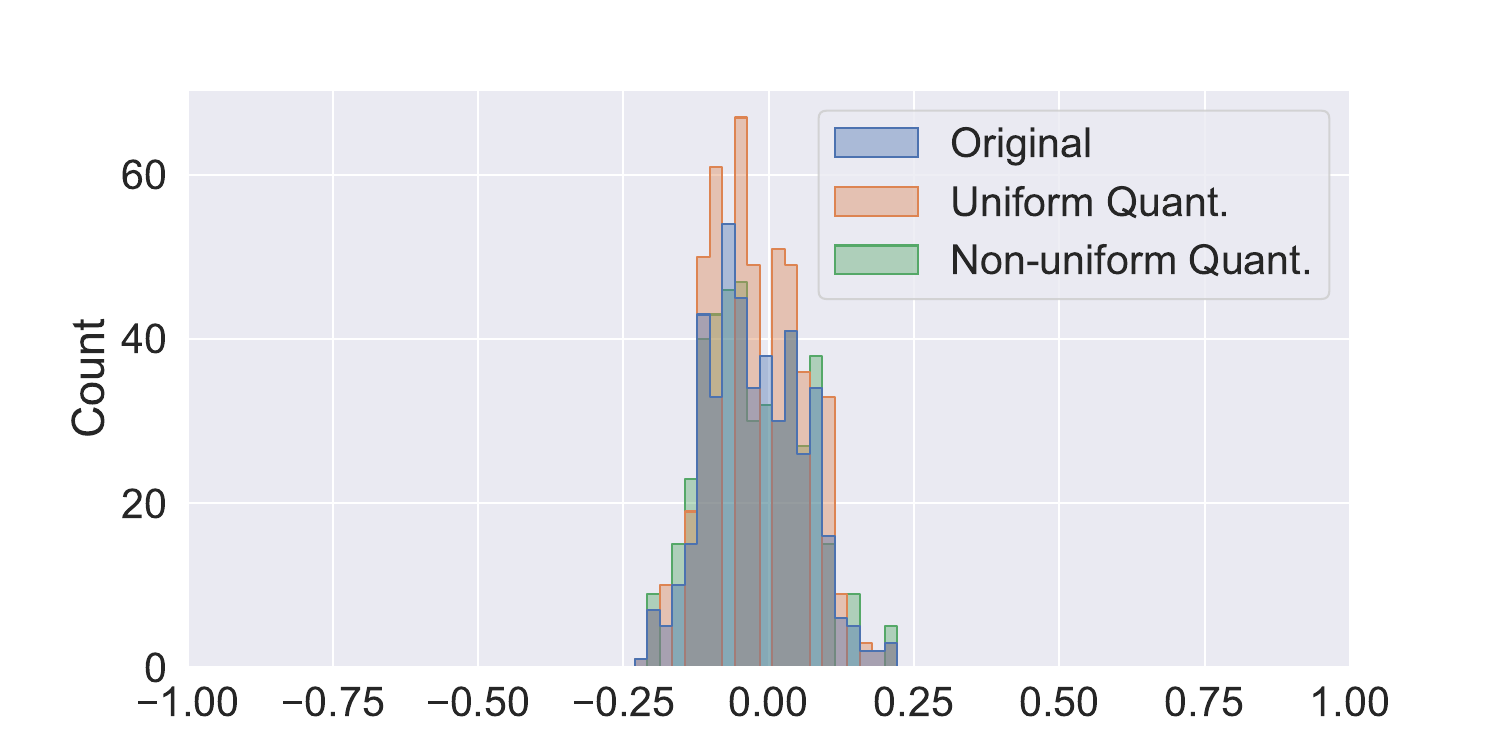}
}\hspace{-12mm} 
\subfigure[Epoch: 4]{
\includegraphics[width=0.28\textwidth]{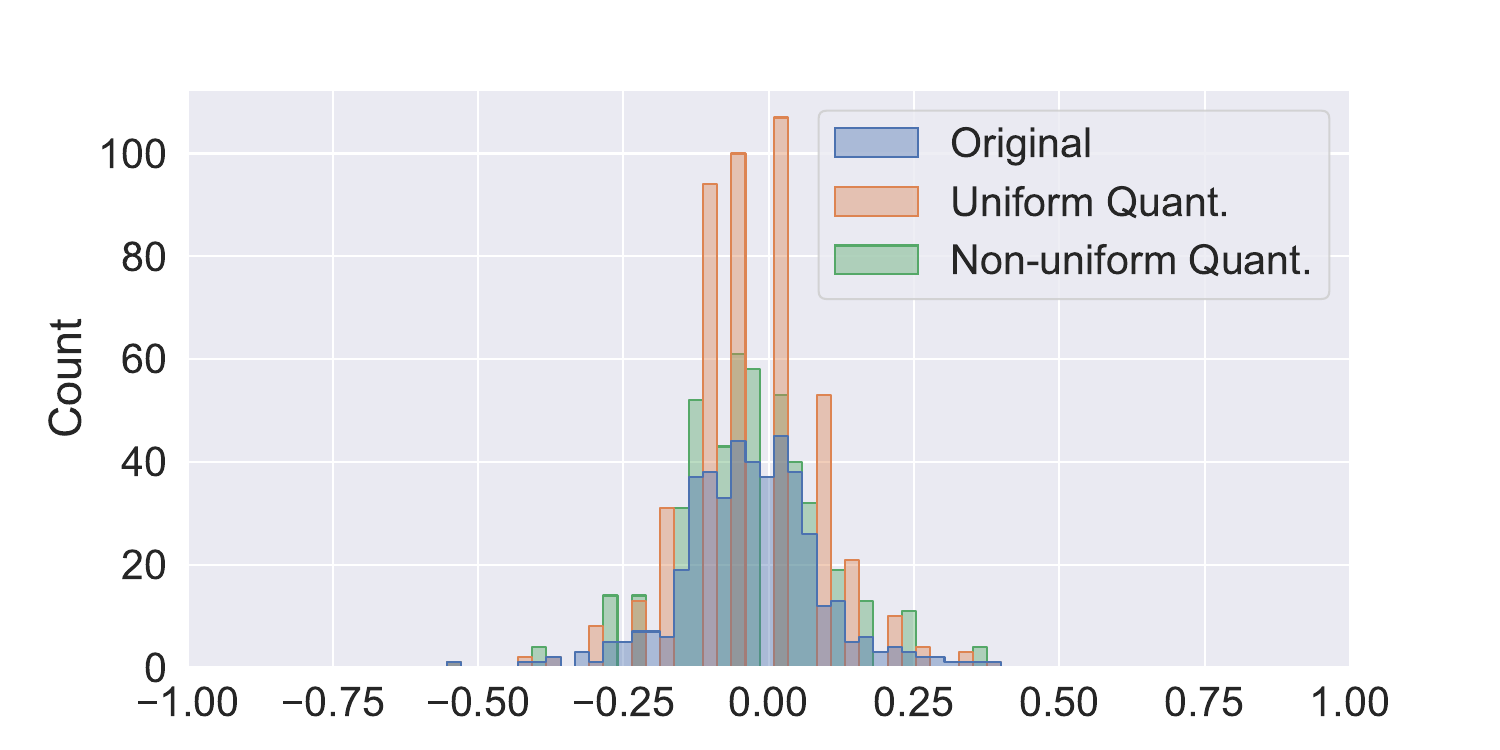}\label{fig_third_case}
}\hspace{-8mm} 
\caption{The effect of 4-bit quantization techniques on the weight distribution over epochs during a single neural network training process.}

\label{fig:quant-dequant-by-method}
\end{figure}

ASYM quantization stands out for its straightforwardness and computational efficiency, with a $O(1)$ complexity. This methodology, however, may not be suitable for values with non-uniform distributions. Since the quantization range is determined by the two extreme values 
($w_{min}$, $w_{max}$), the resulting bins are highly sensitive to these outliers, leading to unnecessarily wide bin widths. As illustrated in Figure~\ref{fig:quant-dequant-by-method}, when the parameter distribution of the neural network follows a normal distribution (bell curve), the majority of quantized values become concentrated near the center of the range. This effect becomes more pronounced as the target bit-width for quantization decreases or as the original weight distribution exhibits greater variance over the training, further amplifying the deviation from the original distribution.

Non-uniform quantization methods employ more sophisticated strategies that consider the underlying distribution of values when mapping them to smaller intervals. In contrast, uniform quantization disregards the density of values, potentially leading to an ``outlier effect" where extreme values disproportionately affect the quantization process. One approach to mitigate this issue is to use K-means clustering for interval determination~\cite{gong2014compressing, wu2016quantized}.

In this context, K-means clustering operates on a one-dimensional dataset (flattened model weights) and determines the number of clusters as $2^\mathcal{N}$, where $\mathcal{N}$ is the number of quantization bits. The algorithm computes $2^\mathcal{N}$ centroids to represent the interval values, effectively clustering the weights based on their distribution. By leveraging K-means clustering, this method reduces the influence of extreme weight values and preserves the original distribution more effectively than uniform quantization, as demonstrated in Figure~\ref{fig:quant-dequant-by-method}. However, the computational demands of K-means clustering are higher compared to uniform quantization. The detailed algorithm for this quantization process is provided in Algorithm~\ref{alg:k-means-quantization}.

\begin{algorithm}[htbp]
\caption{K-means clustering-based Weight Quantization}
\label{alg:k-means-quantization}

\textbf{Quantization (Q):}

\textbf{Input:} Weights $\mathbf{w}$, number of bits $\mathcal{N}$

\textbf{Output:} Quantized weights $\mathbf{Q(w)}$, codebook $\mathcal{B}$

\begin{algorithmic}[1]
\STATE Initialize $k = 2^\mathcal{N}$ clusters for K-means clustering.
\STATE Perform K-means clustering on the weights $\mathbf{w}$ to compute $k$ centroids $\mathcal{B} = \{\mathbf{b}_1, \mathbf{b}_2, \dots, \mathbf{b}_k\}$ and assign each weight $w_i$ to the nearest centroid.
\STATE Replace each weight $w_i \in \mathbf{w}$ with the index of its nearest centroid to obtain the quantized weights $\mathbf{Q(w)}$.
\STATE Return the quantized weights $\mathbf{Q(w)}$ and the codebook $\mathcal{B}$.
\end{algorithmic}

\textbf{Dequantization (DeQ):}

\textbf{Input:} Quantized weights $\mathbf{Q(w)}$, codebook $\mathcal{B}$

\textbf{Output:} Dequantized weights $\tilde{\mathbf{w}}$

\begin{algorithmic}[1]
\STATE Replace each quantized weight index in $\mathbf{Q(w)}$ with its corresponding centroid value in $\mathcal{B}$ to reconstruct the dequantized weights $\tilde{\mathbf{w}}$.
\STATE Return the dequantized weights $\tilde{\mathbf{w}}$.
\end{algorithmic}
\end{algorithm}

\section{Proposed Method: FedShift}
\begin{figure}[!ht]
\centering
\includegraphics[width=0.4\textwidth]{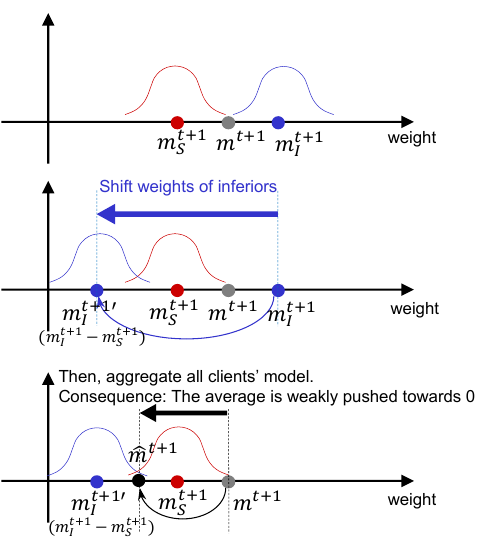}
\caption{Concept visualization of \texttt{FedShift}}
\label{fig:visualization}
\end{figure}
\subsection{\texttt{FedShift} - Weight Shifting Aggregation}\label{proposed-fedshift}

To mitigate performance degradation caused by quantization and client drift without incurring additional transmission or storage overhead, we employ a direct statistical matching approach on the weights. Inspired by transformation-based techniques such as weight standardization and weight normalization~\cite{salimans2016weight, huang2017centered, qiao2019micro}, which have been shown to enhance deep learning training, we focus specifically on weight shifting via mean subtraction. In this approach, we treat the weight distribution of non-quantized models as the reference for the correct distribution and therefore do not modify them. Instead, the shifting technique is applied exclusively to quantized models to correct distributional distortions and restore the desired statistical properties.

First, we need to note that the following equations describe the weight within a layer of the neural network model.
Let the client sets $\mathbb{I}$ and $\mathbb{S}$ denotes the \textit{inferior} and \textit{superior} clients selected at each round, and let $I$ and $S$ be the number of clients in each set, respectively. 
In the aggregation stage, the weights of \textit{inferior} clients $i \in \mathbb{I}$ are shifted by subtracting the average weight $m^{t+1}$. The shifted weight of an \textit{inferior} client $i$ is calculated as follows:

\begin{equation}
{\mathbf{w}_{i}^{t+1}}'=\mathbf{w}_{i}^{t+1} - m^{t+1}\vec{\mathbf{1}}_P , \hspace{5mm} \forall i \in \mathbb{I}
\end{equation}

where $\mathbf{1}_P$ denotes the one vector of size $P$ and the weight average $m^{t+1}$ is the average of the weights of aggregated global model of all clients (\textit{inferior}$+$\textit{superior}) at round $t+1$ before shifting.

\begin{equation}
    m^{t+1} =\frac{1}{P}\sum_p \mathbf{w}^{t+1}[p]
\end{equation}
Here, $p$ is the index of weight parameter, $P$ is the total number of parameters and $\mathbf{w}^{t+1}$ is the aggregated global model before shifting.
The global model of \texttt{FedShift} is updated by the aggregation of \textit{superior} clients and shifted \textit{inferior} clients as follows. 



\begin{equation}
\hat{\mathbf{w}}^{t+1}[p]=\frac{1}{K}\left( \sum_{i\in \mathbb{I}}{ \mathbf{w}_{i}^{t+1}}'[p]+\sum_{j \in \mathbb{S}} \mathbf{w}_{j}^{t+1}[p] \right)
\end{equation}

Ultimately, the shifted global model $\hat{\mathbf{w}}^{t+1}$ can be simplified in the vector notation as follows:

\begin{equation}\label{eq:final-form}
\hat{\mathbf{w}}^{t+1} = \mathbf{w}^{t+1} - \frac{I}{K}m^{t+1}\vec{\mathbf{1}}_P.
\end{equation}


If we examine the result of shifting, it pushes the average weight to 0 with the proportion of $S/K$, which is the ratio of superior clients to all clients.
It can be easily seen by examining the average of shifted weights as follows.
\begin{equation}\label{eq:m}
    \begin{array}{l}
         \displaystyle \hat{m}^{t+1}=\frac{1}{PK}\left(\sum_{p=1}^P \sum_{k=1}^K{ (w_{k}^{t+1}}[p]-\frac{I}{K}m^{t+1})\right)\\
         \displaystyle = m^{t+1}-\frac{I}{K}m^{t+1} = \frac{S}{K}m^{t+1}.
    \end{array}
\end{equation}
Since $\mathbb{S}$ is the subset of $\mathbb{K}$, $0 \leq
|\hat{m}^{t+1}| \leq
 |m^{t+1}|$  holds. Figure~\ref{fig:visualization} provides a conceptual overview of \texttt{FedShift}.

\begin{algorithm}[htbp]
\footnotesize 
\caption{FedShift with DeQ for Inferior Clients}
\label{alg:fed-shift-deq}
\begin{algorithmic}[1]
\STATE \textbf{Input:} Clients \(N\), participation ratio \(C\), local epochs \(E\), communication rounds \(T\), local learning rate \(\eta\), initial parameters \(\mathbf{w}^0\), set of inferior clients \(\mathbb{I}\)
\STATE \textbf{Note:} Parameter \(\mathbb{I}\) represents the group of inferior clients that are pre-identified. 
\FOR{round \(t = 1\) to \(T\)}
    \STATE Randomly Select \(\mathbb{K}\) from  \(C \cdot N\) clients
    \FOR{each client \(k \in \mathbb{K}\) \textbf{in parallel}}
        \STATE \(\mathbf{w}_{k}^t \leftarrow \hat{\mathbf{w}}^{t}\)
 
        \FOR{epoch \(e = 1\) to \(E\)}
            \STATE \(\mathbf{w}_{k}^{t,e+1} \leftarrow \mathbf{w}_{k}^{t,e} - \eta \nabla F_k(\mathbf{w}_{k}^{t,e})\)
        \ENDFOR
        \IF{$k \in \mathbb{I}$}
            \STATE \(\mathbf{w}_{k}^{t+1} \leftarrow Q(\mathbf{w}_{k}^{t+1})\) \COMMENT{Quantization process for inferior clients}
        \ENDIF
    \ENDFOR
    
    \FOR{each inferior client \(k \in \mathbb{I}^t\)}
        \STATE \(\mathbf{w}_{k}^{t+1} \leftarrow DeQ(\mathbf{w}_{k}^{t+1})\) \COMMENT{Restore inferior client updates}
    \ENDFOR
    \vspace{2mm}
    \STATE \(\mathbf{w}^{t+1} \leftarrow \sum_{k=1}^K p_k\mathbf{w}_k^t\) \label{code:agg}
    \vspace{2mm}

    \STATE \(m^{t+1}=\frac{1}{P}\sum_p \left( \mathbf{w}^{t+1}[p]\right)\)\label{code:mean-calc}
    \vspace{1mm}

    \STATE \(\hat{\mathbf{w}}^{t+1} \leftarrow \mathbf{w}^{t+1} - \frac{|\mathbb{I}^t|}{K}m^{t+1}\vec{\mathbf{1}}_P\)\label{code:shifting}

\ENDFOR
\STATE \textbf{Output:} Global parameters \(\mathbf{w}^T\)
\end{algorithmic}
\label{alg:fed-shift}
\end{algorithm}

To summarize, \texttt{FedShift} can be represented as Algorithm~\ref{alg:fed-shift}. The overall process follows the existing FedAvg method. After the server-side weight aggregation (line \ref{code:agg}), the adjustment of global model weights is included, along with additional shifting operations (line~\ref{code:mean-calc}-\ref{code:shifting}).

\subsection{Theoritical Analysis}\label{theoritical-analysis}
\textbf{Convergence Analysis.} Here, we present the convergence analysis of \texttt{FedShift} on non-IID data distribution as the following theorem.
It implies that when the average of weights, $M$, goes to $0$, \texttt{FedShift} algorithm converges to the solution with the convergence rate of $\mathcal{O}(\frac{1}{T})$.
This finding is in accordance with the simulation results in which the average of weights becomes very close to $0$ and shows stable training with \texttt{FedShift}. 

We have the following assumptions on the functions $F_1, \cdots, F_N$ where Assumptions~\ref{AS1} and \ref{AS2} are standard and Assumptions~\ref{AS3} and \ref{AS4} are also typical which have been made by the works \cite{zhang2012communication}, \cite{stich2018local} and \cite{yu2019parallel}.

\begin{assumption}\label{AS1}
\textit{L-Smoothness: For all $\mathbf{w}$ and $\mathbf{w'}$, $F_k(\mathbf{w})\leq F_k(\mathbf{w'})+(\mathbf{w}-\mathbf{w'})^\intercal\nabla F_k(\mathbf{w'})+\frac{L}{2} || \mathbf{w}-\mathbf{w'}||_2^2$}
\end{assumption}

\begin{assumption}\label{AS2}
\textit{$\mu$-strongly convex: For all $\mathbf{w}$ and $\mathbf{w'}$, $F_k(\mathbf{w})\geq F_k(\mathbf{w'})+(\mathbf{w}-\mathbf{w'})^\intercal\nabla F_k(\mathbf{w'})+\frac{\mu}{2} || \mathbf{w}-\mathbf{w'}||_2^2$}
\end{assumption}

\begin{assumption}\label{AS3}
\textit{Variance of gradients are bounded: Let $\xi_t^k$ be the data randomly sampled from k-th client's data. The variance of stochastic gradients in each device is bounded as: $\mathbb{E}[||\nabla F_k(\mathbf{w}_k^\tau, \xi_k^\tau)-\nabla F_k(\mathbf{w}_k^\tau)||_2^2] \leq \sigma_k^2$ for $k=1,\cdots, N$}
\end{assumption}

\begin{assumption}\label{AS4}
\textit{Stochastic gradients are bounded: $\mathbb{E}\left[||\nabla F_k(\mathbf{w}_k^\tau, \xi_k^\tau)||_2^2\right] \leq G^2$ for $k=1,\cdots, N$ and $\tau=1,2,\cdots$ }
\end{assumption}

Here, we have additional assumption regarding the weight distribution of training model. This is a reasonable assumption because FL training would converge at some point then average of model's weights does not diverge to positive/negative infinity unless gradient explodes.

\begin{assumption}\label{AS5}
\textit{Weight average is bounded: $\mathbb{E}\left[(m^\tau)^2\right] \leq M$ for all $\tau$ }
\end{assumption}

\begin{theorem}\label{th1}
    Under the Assumptions \ref{AS1} to \ref{AS5} regarding $L$-smoothness \& $\mu$-convexity of function and bounded gradients and weight average, for the choice of $\beta = \frac{2}{\mu}$, $\gamma = \max\{ \frac{8L}{\mu}, E\}-1$, $\varepsilon > 0$, then $\eta_\tau=\frac{2}{\mu}\frac{1}{\gamma+\tau}$, the error after $\tau$ local steps of update with full client participation satisfies
\begin{align*}
&\mathbb{E}[F(\overline{\mathbf{w}}_\tau) ]- F^* \\
&\leq \underbrace{\frac{L}{2} \frac{1}{\tau+\gamma} \left( \frac{B_1 \beta^2 }{\left(1 + \frac{1}{\varepsilon}\right) \beta \mu - \frac{1}{\varepsilon}\left(\tau+\gamma\right)-1} + (\gamma + 1) \Delta_1 \right) }_{\text{Vanishing term}} \\
&+ \underbrace{\frac{L}{2}\left(\frac{B_2 M}{\left(1+\frac{1}{\varepsilon}\right)\beta\mu\frac{1}{\tau+\gamma}-\frac{1}{\varepsilon}-\frac{1}{\tau+\gamma}}\right)}_{\text{Vanishes when $M\rightarrow 0$}}
\end{align*}

where $B_1$, $B_2$ and $M$ are values determined by bounds, $F^*$ is the minimum point of function $F$, and $\Delta_{\tau=1} =\mathbb{E}\|\overline{\mathbf{w}}^{\tau=1} - \mathbf{w}^*\|^2$.

\begin{proof}
    See Appendix \ref{CAproof} for the detailed proof and notations.
\end{proof}
\end{theorem}


\begin{figure}[!htbp]
\centering
\includegraphics[width=0.45\textwidth]{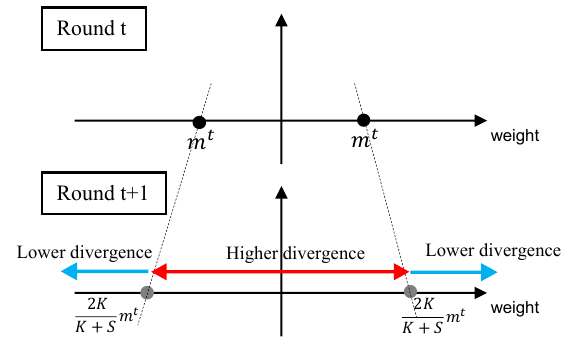}
\caption{Illustration of \texttt{FedShift} balancing divergence}
\label{how-we-control-divergence}
\end{figure}

\textbf{Divergence Analysis.}\label{div-analyis}
In another aspect, we provide an analysis of the divergence of the model between the rounds. 
We examine how much the global model diverges from the previous round by comparing it with the baseline algorithm \texttt{FedAvg}.
The model divergence from the previous round to the current round is considered one of the measures of stable FL training \cite{rehman2023dawa}. 
If divergence is too large between rounds, it may ruin the collaboration between clients, but too small divergence would result in slow training. \texttt{FedShift} balances the divergence between rounds based on the average weight of the previous and current rounds, i.e., $m^t$, $m^{t+1}$.
Figure \ref{how-we-control-divergence}, shows that depending on the  $m^t:m^{t+1}$ ratio, the divergence differs.
The result is summarized in the following theorem.
We encourage the readers to refer to the appendix for detailed proof and interpretations with examples.

\begin{theorem}\label{th2}
    Let $D$ be the L2 divergence from the previous round global model to the current round global model. Then, the divergence of \texttt{FedShift} and \texttt{Fedavg} satisfies the following condition

    \begin{align}
    \displaystyle D_{FA}^2-D_{FS}^2 &= \frac{IP}{K}\left(\frac{K+S}{K}(m^{t+1})^2 -2\frac{m^{t}}{(m^{t+1})^2}\right)
    \end{align}
\begin{proof}
    See Appendix \ref{DAproof} for the detailed proof and interpretations.
\end{proof}
\end{theorem}

\section{Experiment and Result}\label{experiment}

This section describes our experiment, including the dataset, relevant models, and FL environment settings. Then, performance evaluations of our proposed \texttt{FedShift} algorithm and other conventional algorithms are presented in various aspects.

\subsection{Experimental Setup}\label{dataset}

\subsubsection{Model Architecture}
Our backbone model follows a Convolutional Neural Network (CNN) structure similar to the one presented in \texttt{FedAvg}~\cite{mcmahan2017communication}. It contains two convolutional layers with max-pooling, followed by two fully connected layers. Network weights are initialized via Kaiming initialization~\cite{he2015delving}.

\subsubsection{Dataset and Data Partitioning}

\begin{figure}[htbp]
\centering
\subfigure[\textit{Superior} group]{
    \begin{tikzpicture}[scale=0.48, transform shape]
        \begin{axis}[
        xbar stacked,   
        xmin=0,         
        ytick=data,     
        legend style={at={(axis cs:65,0.2)},anchor=south west},
        yticklabels from table={\supdata}{Label}  
        ]
        \addplot [fill=green!80] table [x=Class0, meta=Label,y expr=\coordindex] {\supdata};   
        \addplot [fill=blue!60] table [x=Class2, meta=Label,y expr=\coordindex] {\supdata};
        \addplot [fill=orange!60] table [x=Class4, meta=Label,y expr=\coordindex] {\supdata};
        \addplot [fill=gray!60] table [x=Class6, meta=Label,y expr=\coordindex] {\supdata};
        \addplot [fill=red!60] table [x=Class8, meta=Label,y expr=\coordindex] {\supdata};
        \legend{Class0, Class2, Class4, Class6, Class8}
        \end{axis}
    \end{tikzpicture}
}
\subfigure[\textit{Inferior} group]{
   \begin{tikzpicture}[scale=0.48, transform shape]
        \begin{axis}[
        xbar stacked,   
        xmin=0,         
        ytick=data,     
        legend style={at={(axis cs:65,0.2)},anchor=south west},
        yticklabels from table={\infdata}{Label}  
        ]
        \addplot [fill=yellow!80] table [x=Class1, meta=Label,y expr=\coordindex] {\infdata};   
        \addplot [fill=purple!60] table [x=Class3, meta=Label,y expr=\coordindex] {\infdata};
        \addplot [fill=pink!60] table [x=Class5, meta=Label,y expr=\coordindex] {\infdata};
        \addplot [fill=black!60] table [x=Class7, meta=Label,y expr=\coordindex] {\infdata};
        \addplot [fill=cyan!60] table [x=Class9, meta=Label,y expr=\coordindex] {\infdata};
        \legend{Class1, Class3, Class5, Class7, Class9}
        \end{axis}
    \end{tikzpicture}
}
\caption{Data partitioning with \textit{inferior} and \textit{superior} group}

\label{fig:data_partition}

\end{figure}
We use the CIFAR-10 dataset~\cite{cifar10}, which has 10 classes, 50{,}000 training images, and 10{,}000 testing images. For experimentation, the training set is distributed among the clients, while the test set remains centralized on the server for evaluating global model performance in terms of top-1 accuracy. To highlight the effects of mixed quantization, we devise a custom shard-based partitioning. We divide CIFAR-10’s classes into two groups without overlapping labels: the \textit{superior} group (even-numbered labels) and the \textit{inferior} group (odd-numbered labels), as shown in Figure~\ref{fig:data_partition}. Each client is assigned at most two labels from its respective group, ensuring a controlled distribution of data.

\subsubsection{Training Configuration}
We consider a federated learning (FL) setup with $N = 100$ clients. At each global round, a subset of clients is randomly selected based on a client participation ratio $C = 0.1$. We run a total of $T = 800$ training rounds. Each selected client trains for $E = 10$ local epochs with a batch size of 50. We employ stochastic gradient descent (SGD) with a learning rate of 0.005 and momentum of 0.9.

\subsubsection{Quantization Setting}
To study the impact of quantization on model performance, we use five different bit widths (4 to 8 bits) for weights in the \textit{inferior} group, while keeping 32-bit precision for the \textit{superior} group. This yields five distinct mixed-precision configurations. For our primary experiment, we apply layer-wise non-uniform (Algorithm~\ref{alg:k-means-quantization}) quantization. Additional results using uniform quantization and Dirichlet-distributed datasets are provided in later sections.

\subsubsection{Baseline FL Methods}
We evaluate the effectiveness of \texttt{FedShift} by integrating it with four baseline FL algorithms: \texttt{FedProx}~\cite{li2020fedprox}, \texttt{SCAFFOLD}~\cite{karimireddy2020scaffold}, \texttt{FedAvg}~\cite{reisizadeh2020fedpaq}, and \texttt{Fed-EF}~\cite{li2023analysis}. In \texttt{Fed-EF}'s original form, it compresses and transmits gradients; for our experiments, we adapt these methods to quantize and transmit weights instead.

\subsubsection{Evaluation Protocol}
We evaluate the global model after each communication round using top-1 accuracy on the central test set. Given the inherent randomness in federated learning, each experiment is repeated three times with different random seeds, and the results are averaged.

\subsection{Experimental Result}\label{evaluation}

\begin{table*}
\small
\caption{Performance summary of FL algorithms under various mixed-precision settings, with and without the integration of \texttt{FedShift}}
\centering
\setlength{\tabcolsep}{0.55em} 

\begin{tabular}{|c|c|c|c|c|c|c|c|c|c|c|}
\hline
Bits & \multicolumn{2}{c|}{32+4} & \multicolumn{2}{c|}{32+5} & \multicolumn{2}{c|}{32+6} & \multicolumn{2}{c|}{32+7} & \multicolumn{2}{c|}{32+8} \\
\hline
Model & Baseline & FedShift & Baseline & FedShift & Baseline & FedShift & Baseline & FedShift & Baseline & FedShift \\
\hline
\texttt{FedPAQ} & 59.7\textcolor{gray}{$\pm$1.3} & \cellcolor{yellow!25}\textbf{62.5\textcolor{gray}{$\pm$1.3}} & 62.7\textcolor{gray}{$\pm$1.1} & \cellcolor{yellow!25}\textbf{66.9\textcolor{gray}{$\pm$0.9}} & 64.3\textcolor{gray}{$\pm$0.9} & \cellcolor{yellow!25}\textbf{69.3\textcolor{gray}{$\pm$1.1}} & 65.4\textcolor{gray}{$\pm$1.0} & \cellcolor{yellow!25}\textbf{70.2\textcolor{gray}{$\pm$0.8}} & 66.0\textcolor{gray}{$\pm$0.9} & \cellcolor{yellow!25}\textbf{70.4\textcolor{gray}{$\pm$0.8}} \\
\texttt{FedEF} & 64.8\textcolor{gray}{$\pm$1.6} & \cellcolor{yellow!25}\textbf{65.5\textcolor{gray}{$\pm$3.6}} & 65.3\textcolor{gray}{$\pm$1.7} & \cellcolor{yellow!25}\textbf{67.0\textcolor{gray}{$\pm$0.5}} & 64.9\textcolor{gray}{$\pm$1.8} & \cellcolor{yellow!25}\textbf{68.8\textcolor{gray}{$\pm$1.0}} & 64.9\textcolor{gray}{$\pm$2.6} & \cellcolor{yellow!25}\textbf{70.3\textcolor{gray}{$\pm$0.6}} & 65.5\textcolor{gray}{$\pm$1.9} & \cellcolor{yellow!25}\textbf{70.1\textcolor{gray}{$\pm$0.7}} \\
\texttt{FedProx} & 60.1\textcolor{gray}{$\pm$1.3} & \cellcolor{yellow!25}\textbf{62.6\textcolor{gray}{$\pm$1.2}} & 63.5\textcolor{gray}{$\pm$0.9} & \cellcolor{yellow!25}\textbf{66.6\textcolor{gray}{$\pm$1.0}} & 65.2\textcolor{gray}{$\pm$0.9} & \cellcolor{yellow!25}\textbf{69.0\textcolor{gray}{$\pm$0.7}} & 66.4\textcolor{gray}{$\pm$0.7} & \cellcolor{yellow!25}\textbf{70.5\textcolor{gray}{$\pm$0.6}} & 66.6\textcolor{gray}{$\pm$0.6} & \cellcolor{yellow!25}\textbf{70.5\textcolor{gray}{$\pm$0.6}} \\
\texttt{SCAFFOLD} & 62.9\textcolor{gray}{$\pm$2.0} & \cellcolor{yellow!25}\textbf{65.9\textcolor{gray}{$\pm$1.5}} & 66.5\textcolor{gray}{$\pm$2.7} & \cellcolor{yellow!25}\textbf{67.9\textcolor{gray}{$\pm$1.6}} & 66.9\textcolor{gray}{$\pm$1.4} & \cellcolor{yellow!25}\textbf{68.8\textcolor{gray}{$\pm$1.3}} & 66.7\textcolor{gray}{$\pm$1.8} & \cellcolor{yellow!25}\textbf{69.7\textcolor{gray}{$\pm$1.5}} & 66.8\textcolor{gray}{$\pm$1.7} & \cellcolor{yellow!25}\textbf{69.6\textcolor{gray}{$\pm$1.5}} \\
\hline
\end{tabular}

\label{tab:final_acc_result}

\end{table*}
\begin{figure*}[htbp]
\centering
\includegraphics[width=\textwidth]{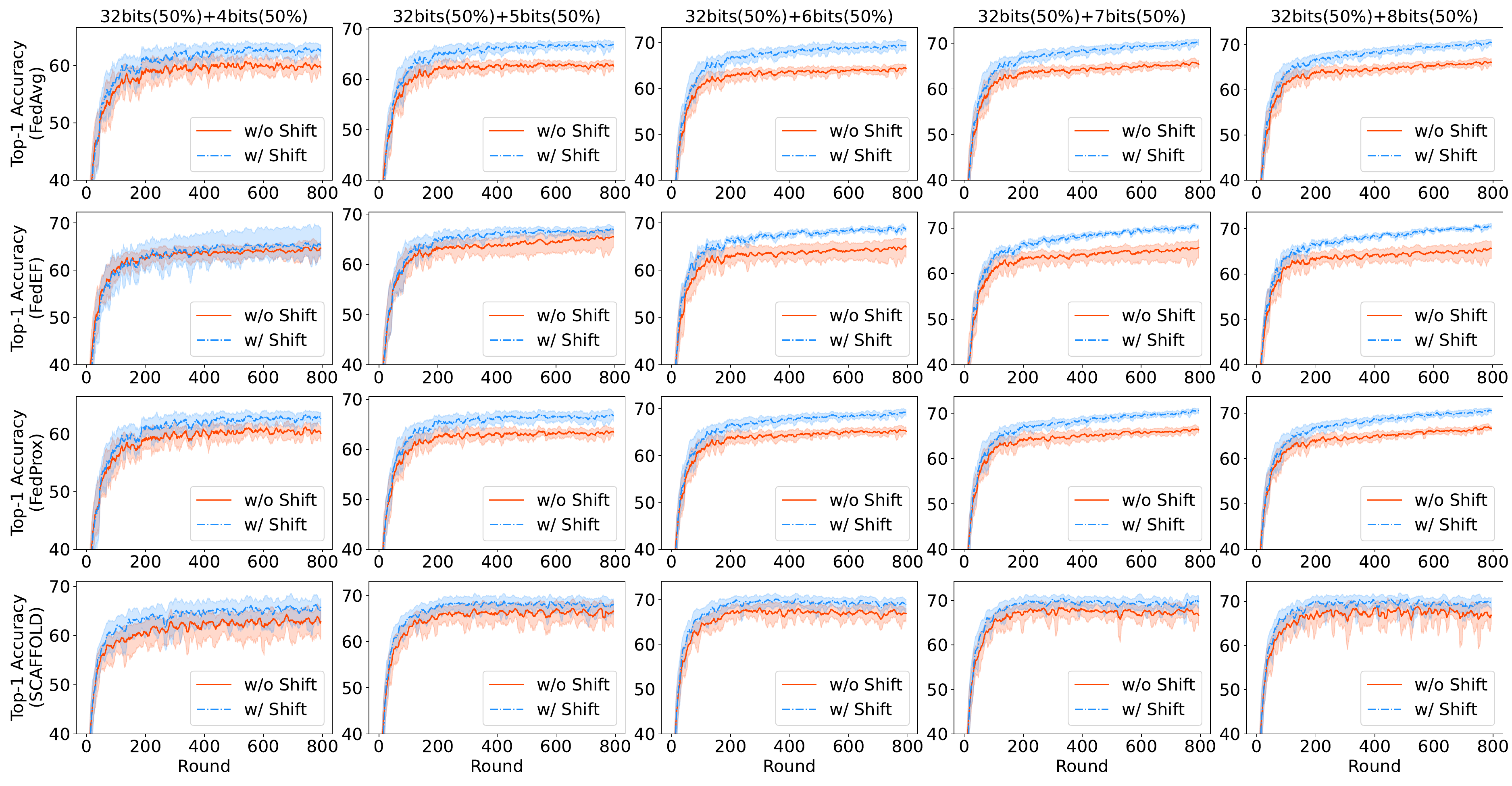}
\caption{Training dynamics comparison of four FL algorithms across various bit widths, with and without shift aggregation. Each graph represents the top-1 accuracy smoothed using a simple moving average with a window of 10 rounds. The results are averaged over three independent runs with different seeds. The experiments utilize 50\% of 32-bit width combined with 50\% of 4, 5, 6, 7, and 8-bit widths through non-uniform quantization.}
\label{acc-result-line}
\end{figure*}

\subsubsection{Main Result} 
The main results of the experiments are presented in Table~\ref{tab:final_acc_result} and Figure~\ref{acc-result-line}. First, regarding the impact of quantization bit-width, the performance decreases as the bit-width reduces, regardless of the baseline algorithm. Conversely, as the bit-width increases, the performance gradually converges to a stable level. Although not explicitly shown in the table, the performance of \texttt{FedAvg} with full precision (32+32 bits) was 66.20\%.

Second, analyzing performance across different algorithms, \texttt{FedAvg}, which does not account for quantization errors or client drift, consistently exhibited the lowest performance. For combinations involving 4-bit quantization, \texttt{Fed-EF} demonstrated the best performance, while for other bit-widths, \texttt{SCAFFOLD} showed superior results. Interestingly, at 8-bit quantization, all algorithms achieved comparable performance.

Finally, the integration of \texttt{FedShift} improved the performance of all baseline algorithms across all quantization bit-widths. Notably, even with 5-bit quantization, the combination of \texttt{FedShift} exceeded the performance of full-precision \texttt{FedAvg} (66.20\%), demonstrating that \texttt{FedShift} effectively mitigates both quantization-induced errors and client drift-related performance degradation.

\begin{figure}[!htbp]
    \centering
    \includegraphics[width=0.4\textwidth]{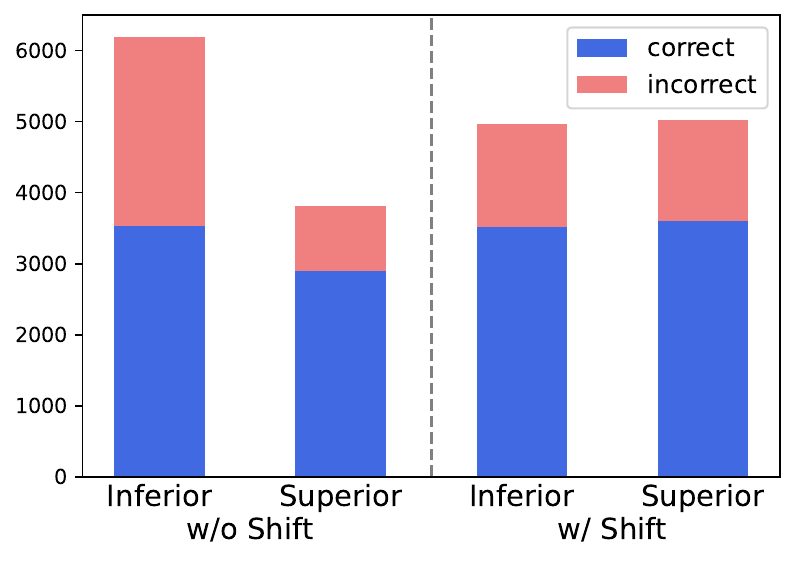}

    \caption{Predicted label distribution comparison between non-shifted and shifted models, where the global model is aggregated with mixed precision.}
    \label{fig:prediction-ratio}
\end{figure}
\subsubsection{Bias Reduction} To understand the performance degradation of the global model under Mixed Precision aggregation and the evolution of its predictions during inference, we analyze the effect of integrating \texttt{FedShift}. In the experimental setup, FL is performed with label quantization applied to a subset of clients. As illustrated in Figure~\ref{fig:prediction-ratio}, the global model’s final predictions exhibit a notable bias toward labels prevalent in the quantized clients. However, when \texttt{FedShift} is applied, the model’s predictions become more uniformly distributed across both quantized and non-quantized groups.

Importantly, when data quantity and the participation of superior and inferior groups per round are held constant, mixed precision aggregation alone amplifies biased predictions, adversely affecting the model’s overall performance. These findings underscore the critical need to mitigate bias in FL systems, particularly under mixed precision conditions.

\subsubsection{Divergence and Client-drift Reduction} 
\begin{figure}[htbp]
    \centering
    \begin{minipage}[b]{0.48\textwidth}
        \centering
        \includegraphics[width=\textwidth]{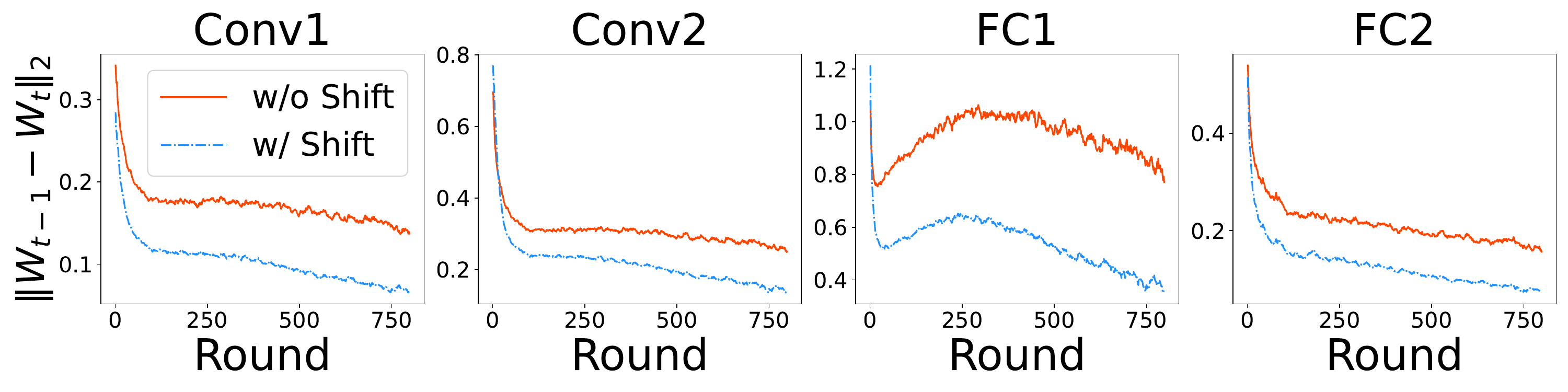}
        
        \label{fig:image1}
    \end{minipage}
    
    \begin{minipage}[b]{0.48\textwidth}
        \centering
        \includegraphics[width=\textwidth]{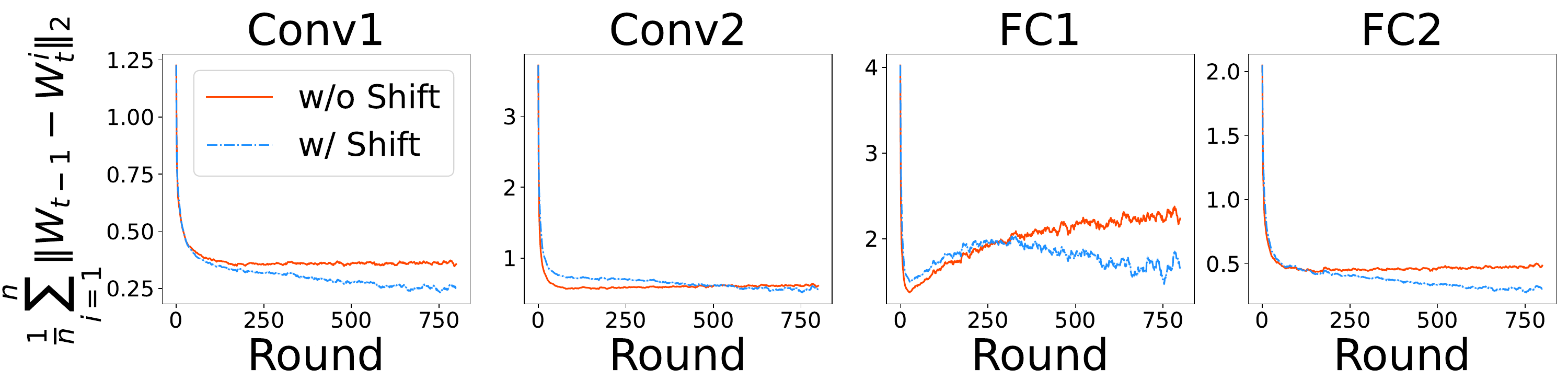}
        
        \label{fig:image2}
    \end{minipage}
    \caption{\textbf{Layer-wise L2 distance measurements.} (Top) The L2 distance between the global model at round $t$ and $t-1$. (Bottom) The average of L2 distances between the global model at round $t-1$ and the selected local models at round $t$.}
    \label{fig:model-drift}
\end{figure}

We examined the consistent improvements in model performance through the lens of round-wise and client-wise divergence. Figure~\ref{fig:model-drift} presents two layer-wise L2 distance metrics, each capturing different aspects of client drift. Each line corresponds to the same model configuration, with (blue) and without (red) the integration of \texttt{FedShift}.

The top plot shows the L2 distance between the global model at rounds $t$ and $t-1$, aligning with the divergence analysis in Theorem~\ref{th2}. Under conditions of low client participation rate ($C$), there is a higher likelihood of selecting different clients in each round. This variation exposes the global model to disparate datasets across rounds, resulting in unstable training dynamics and larger round-to-round divergence. Notably, as shown in the plot, \texttt{FedShift} effectively reduces the global model divergence across consecutive rounds compared to models without shift adjustments.

The bottom plot depicts the average L2 distance between the global model at round $t-1$ and the selected local models at round $t$, representing the degree of client drift within a given round. Higher client drift corresponds to greater distances between the initial global model at $t-1$ and the locally trained models at $t$. Again, the application of \texttt{FedShift} significantly mitigates the client drift effect, offering more stable convergence compared to non-shifting algorithms.


\begin{table*}
\small
\caption{Top-1 Accuracy for Varying Dirichlet Alpha Using Different Quantization Methods, Measured After 800 Rounds and Averaged Over Three Independent Runs with Different Random Seeds.}
\centering
\setlength{\tabcolsep}{0.35em} 
\begin{tabular}{|c|c|c|c|c|c|c|c|c|c|c|c|}

\hline
\multirow{2}{*}{Dirichlet} &  \multirow{2}{*}{Shifting} & \multicolumn{10}{c|}{Uniform Quantization} \\
\cline{3-12}
&& \multicolumn{2}{c|}{32+4bit} & \multicolumn{2}{c|}{32+5bit} & \multicolumn{2}{c|}{32+6bit} & \multicolumn{2}{c|}{32+7bit} & \multicolumn{2}{c|}{32+8bit} \\
\hline
\multirow{2}{*}{0.1} &  w/o Shift & N/A &\multirow{2}{*}{-} & 45.6\textcolor{gray}{$\pm$6.79}&\multirow{2}{*}{\textcolor{red}{+12.7}}& 59.0\textcolor{gray}{$\pm$1.83}&\multirow{2}{*}{\textcolor{red}{+2.5}}& 62.5\textcolor{gray}{$\pm$1.35}&\multirow{2}{*}{\textcolor{red}{+1.8}}& 62.5\textcolor{gray}{$\pm$1.64}&\multirow{2}{*}{\textcolor{red}{+2.2}}\\
& w/ Shift & N/A && \textbf{58.3}\textcolor{gray}{$\pm$2.70}&& \textbf{61.5}\textcolor{gray}{$\pm$1.91}&& \textbf{64.3}\textcolor{gray}{$\pm$1.35}&& \textbf{64.7}\textcolor{gray}{$\pm$1.58}& \\
\hline
\multirow{2}{*}{0.5} &  w/o Shift & N/A &\multirow{2}{*}{-}& 43.2\textcolor{gray}{$\pm$28.63}&\multirow{2}{*}{\textcolor{red}{+16.2}}& 67.3\textcolor{gray}{$\pm$1.11}&\multirow{2}{*}{\textcolor{red}{+2.8}}& 68.9\textcolor{gray}{$\pm$0.60}&\multirow{2}{*}{\textcolor{red}{+2.6}}& 68.8\textcolor{gray}{$\pm$0.94}&\multirow{2}{*}{\textcolor{red}{+3.3}} \\
& w/ Shift & N/A& & \textbf{59.4}\textcolor{gray}{$\pm$2.27}&& \textbf{70.1}\textcolor{gray}{$\pm$0.98}&& \textbf{71.5}\textcolor{gray}{$\pm$0.82}&& \textbf{72.1}\textcolor{gray}{$\pm$0.71}& \\
\hline

\multirow{2}{*}{100} &  w/o Shift & N/A&\multirow{2}{*}{-} & 64.3\textcolor{gray}{$\pm$1.97}&\multirow{2}{*}{\textcolor{red}{+3.1}}& 71.1\textcolor{gray}{$\pm$0.46}&\multirow{2}{*}{\textcolor{red}{+3.0}}& 72.2\textcolor{gray}{$\pm$0.49}&\multirow{2}{*}{\textcolor{red}{+3.3}}& 72.1\textcolor{gray}{$\pm$0.12}&\multirow{2}{*}{\textcolor{red}{+3.4}}\\
& w/ Shift & N/A& & \textbf{67.4}\textcolor{gray}{$\pm$2.89}& & \textbf{74.1}\textcolor{gray}{$\pm$0.40}& & \textbf{75.5}\textcolor{gray}{$\pm$0.85}& & \textbf{75.5}\textcolor{gray}{$\pm$0.57}& \\
\hline
\hline
\multirow{2}{*}{Dirichlet} & \multirow{2}{*}{Shifting} & \multicolumn{10}{c|}{Non-uniform Quantization} \\
\cline{3-12}
&& \multicolumn{2}{c|}{32+4bit} & \multicolumn{2}{c|}{32+5bit} & \multicolumn{2}{c|}{32+6bit} & \multicolumn{2}{c|}{32+7bit} & \multicolumn{2}{c|}{32+8bit} \\
\hline
\multirow{2}{*}{0.1} & w/o Shift& 53.3\textcolor{gray}{$\pm$1.17}&\multirow{2}{*}{\textcolor{black}{0.0}}& 57.3\textcolor{gray}{$\pm$1.86}&\multirow{2}{*}{\textcolor{red}{+0.5}}& 60.1\textcolor{gray}{$\pm$1.32}&\multirow{2}{*}{\textcolor{red}{+1.4}} & 61.5\textcolor{gray}{$\pm$1.46}&\multirow{2}{*}{\textcolor{red}{+2.2}}& 62.1\textcolor{gray}{$\pm$1.32}&\multirow{2}{*}{\textcolor{red}{+2.0}}\\
& w/ Shift & \textbf{53.3}\textcolor{gray}{$\pm$2.31}&& \textbf{57.8}\textcolor{gray}{$\pm$1.83}& & \textbf{61.5}\textcolor{gray}{$\pm$1.36}& & \textbf{63.7}\textcolor{gray}{$\pm$1.11}& & \textbf{64.1}\textcolor{gray}{$\pm$1.24}& \\
\hline
\multirow{2}{*}{0.5}  & w/o Shift& 63.3\textcolor{gray}{$\pm$1.06}&\multirow{2}{*}{\textcolor{red}{+2.9}}& 65.7\textcolor{gray}{$\pm$1.00}&\multirow{2}{*}{\textcolor{red}{+0.9}}& 66.9\textcolor{gray}{$\pm$1.16}&\multirow{2}{*}{\textcolor{red}{+0.1}} & 68.1\textcolor{gray}{$\pm$0.66}&\multirow{2}{*}{\textcolor{red}{+3.3}} & 66.1\textcolor{gray}{$\pm$0.96}&\multirow{2}{*}{\textcolor{red}{+5.6}}\\
& w/ Shift & \textbf{65.2}\textcolor{gray}{$\pm$1.14}&& \textbf{66.6}\textcolor{gray}{$\pm$0.97}&& \textbf{70.0}\textcolor{gray}{$\pm$0.65}&& \textbf{71.4}\textcolor{gray}{$\pm$0.43}&& \textbf{71.7}\textcolor{gray}{$\pm$0.69}& \\
\hline
\multirow{2}{*}{100}  & w/o Shift & 68.5\textcolor{gray}{$\pm$0.39}&\multirow{2}{*}{\textcolor{red}{+2.1}}& 70.2\textcolor{gray}{$\pm$0.26}&\multirow{2}{*}{\textcolor{red}{+2.1}}& 71.1\textcolor{gray}{$\pm$0.87}&\multirow{2}{*}{\textcolor{red}{+3.1}}& 71.7\textcolor{gray}{$\pm$0.29}&\multirow{2}{*}{\textcolor{red}{+3.5}}& 71.9\textcolor{gray}{$\pm$0.25}&\multirow{2}{*}{\textcolor{red}{+3.6}}\\
& w/ Shift  & \textbf{70.6}\textcolor{gray}{$\pm$0.58}& & \textbf{72.3}\textcolor{gray}{$\pm$0.38}& & \textbf{74.2}\textcolor{gray}{$\pm$0.26}& & \textbf{75.2}\textcolor{gray}{$\pm$0.32}& & \textbf{75.5}\textcolor{gray}{$\pm$0.51}& \\
\hline

\end{tabular}

\label{tab:final_acc_result-dirich-cifar10}
\end{table*}
\subsubsection{Additional experiments}

\begin{figure}[htbp]
    \centering
    \includegraphics[width=\linewidth]{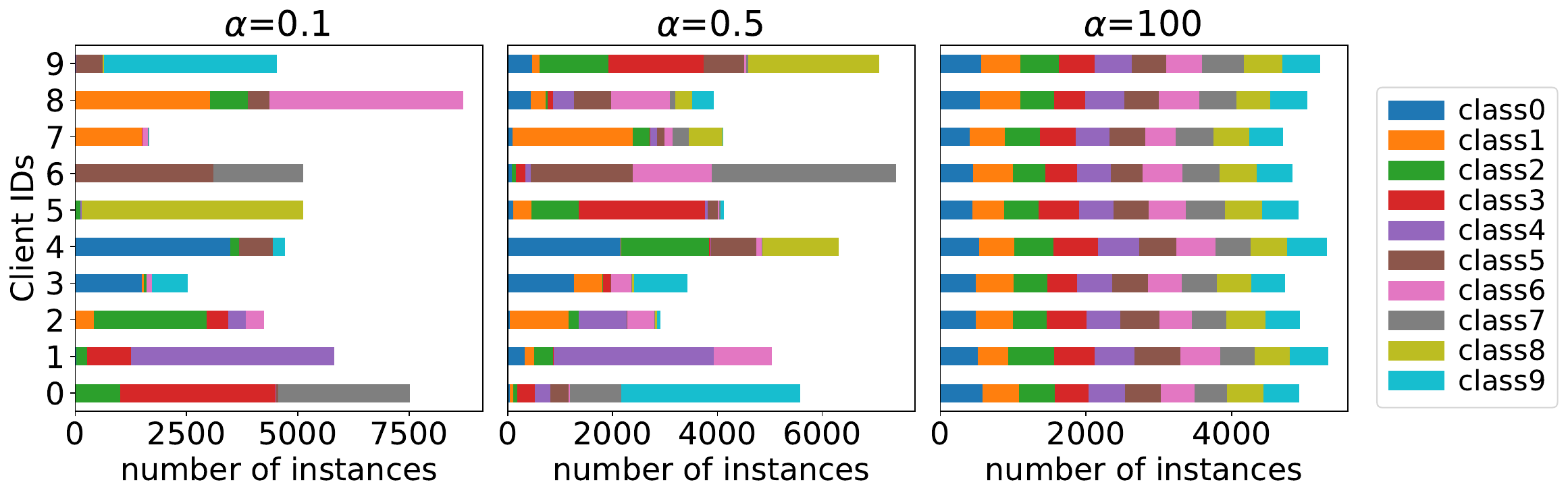}
    \caption{CIFAR10 data partitioning with Dirichlet distribution over 10 clients}
    \label{cifar10-dirichelt}
\end{figure}

We further validate the effectiveness of \texttt{FedShift} using experiments with Dirichlet-distributed data, a commonly adopted method in federated learning research. As shown in Figure~\ref{cifar10-dirichelt}, 100 clients are assigned data partitioned by Dirichlet distribution with $\alpha$ values of 0.1, 0.5, and 100. Half of the clients' local models are quantized (\textit{inferior}), while the other half sends full-precision models (\textit{superior}). Unlike the main experiments, this setup allowed for overlapping labels between \textit{superior} and \textit{inferior} clients, and the data distribution across clients is imbalanced.

Table~\ref{tab:final_acc_result-dirich-cifar10} reports the performance under both quantization methods. Similar to the main experiments, lower bit-widths resulted in degraded performance. Notably, with uniform quantization, training failed in the 4-bit configuration due to interval calculations that do not account for the data distribution. However, from 6 bits onward, uniform quantization exhibited performance comparable to or better than non-uniform quantization. This highlights that, considering the computational overhead of uniform and non-uniform quantization, selecting an appropriate quantization method based on bit-width is a critical trade-off for balancing training speed and overall performance.

Finally, we observe significant performance improvements across all quantization methods, bit-widths, and levels of non-IIDness when applying the shifting technique. This result reinforces the robustness of \texttt{FedShift} in addressing the challenges of FL under diverse settings.





\begin{table}[htbp]
  \centering
  \small
  \caption{Data size and quantization overhead}
\begin{tabular}{|c|c|c|c|c|}
\hline
\multirow{2}{*}{Bits}&Weight&\multirow{2}{*}{Quantization}&Auxiliary&Computing\\
&Size&&Data Size&Overhead\\
\hline
\multirow{2}{*}{4bit} &\multirow{2}{*}{1,052KB}&Uniform&0.25KB&10ms\\
\cline{3-5}
&&Non-uniform&6.19KB&179ms\\
\hline
\multirow{2}{*}{8bit}&\multirow{2}{*}{2,105KB}&Uniform&0.25KB&10ms\\
\cline{3-5}
&&Non-uniform&88.82KB&2,172ms\\
\hline
\hline
32bit&8,421KB&\multicolumn{3}{c|}{Apple M1 Pro, 8 Core, 16GB RAM}\\
\hline
\end{tabular}

\label{table:cost-comparison}

\end{table}

\subsubsection{Quantization Overhead} Lastly, we examine the storage and computational overhead introduced by quantization. Table~\ref{table:cost-comparison} provides a detailed comparison of various bit widths (4-bit, 8-bit, and 32-bit) regarding parameter size in kilobytes (KB), the size of auxiliary data required for dequantization in KB, and the computational overhead for quantization from each client in milliseconds (ms) based on different quantization methods. When evaluating the model parameter sizes, we observe a reduction in total size as the bit width decreases. Specifically, with uniform quantization, only the 32-bit maximum and minimum values need to be transmitted for dequantization, adding a negligible 0.25 KB. Conversely, non-uniform quantization necessitates transmitting each 32-bit centroid, resulting in an increase in size proportional to the number of bits. Considering factors such as training stability, data transmission volume, and computational overhead, we recommend uniform quantization with 8-bit precision for clients with strong network environments. For clients requiring lower bit widths, such as 4-bit quantization, non-uniform quantization becomes a viable option. To assess the efficiency of 4-bit non-uniform quantization compared to 32-bit quantization in terms of communication, we analyze weight transfer time and computational overhead at a network speed of 1 MB/s. With a total data size (including auxiliary data) of 1,059 KB and a computational delay of 179 ms for 4-bit quantization, we establish an efficiency threshold ratio of approximately 7.78. This ratio indicates that 4-bit quantization remains more efficient as long as the network speed is less than 7.78 times slower than that required for 32-bit quantization.

\section{Conclusion}
In this work, we introduce \texttt{FedShift}, a novel algorithm designed to mitigate performance degradation in federated learning caused by mixed-precision aggregation. By leveraging a weight-shifting technique, \texttt{FedShift} addresses key challenges such as global model divergence, client drift, and label bias introduced by quantized client models. Our empirical results demonstrate that \texttt{FedShift} significantly outperforms existing methods by effectively reducing these biases and improving overall model robustness.

\appendices
\section{Proof of the Convergence Analysis}\label{CAproof}

To show that the proposed \texttt{FedShift} algorithm converges with the convergence rate of $\mathcal{O}(\frac{1}{T})$, we provide convergence analysis on \texttt{FedShift} in full participation scenario.
It should be noted that we follow the convergence analysis techniques that is used by \cite{li2019convergence} which is also motivated by many existing literature on convergence analysis, i.e., local SGD convergence \cite{stich2018local}, FedProx \cite{li2018federated}.
Due to the nature of distributed clients, we consider that the data is distibuted non-IID among clients.

\subsection{Problem formulation}
The local objective function of the client $k$ is denoted as $F_k(\mathbf{w})$ and the baseline \texttt{FedAvg} algorithm has the global objective function as

\begin{equation}
    \min _{\mathbf{w}} F(\mathbf{w}) \triangleq \sum_{k=1}^N p_k F_k(\mathbf{w}).
\end{equation}
The optimum value of $F$ and $F_k$ are denoted as $F^*$ and $F_k^*$ and corresponding optimum weights are denoted as $\mathbf{w}^*$ and $\mathbf{w}_k^*$respectively.
Also, $p_k$ is the aggregation weight of the client $k$ where $\sum_{k=1}^{N} p_k = 1$,
since we consider the full participation for the convergence analysis.
Otherwise, the sum of participating clients' aggregation weights at each round should be 1.
In order to analyze the model parameter update, we break down into steps within a round. 
The update of local weight at step $\tau$ can be expressed as follows.

\begin{equation}
    \mathbf{w}_k^{\tau+1} = \mathbf{w}_k^\tau - \eta_\tau \nabla F_k(\mathbf{w}_k^\tau , \xi_k^\tau) 
\end{equation}

For the sequence $\tau=0,1,2,\dots$ and the global synchronization step, $\mathcal{I}_E=\{ nE|n=1,2, \cdots \}$, the weights are uploaded to server only when $\tau+1 \in \mathcal{I}_E$.
For convenience, we let $\overline{\mathbf{w}}^\tau=\sum_{k=1}^N p_k \mathbf{w}_k^\tau $, $\mathbf{g}_\tau = \sum_{k=1}^N p_k \nabla F_k (\mathbf{w}_k^\tau, \xi_k^\tau)$ and
$\overline{\mathbf{g}}_\tau=\sum_{k=1}^N p_k \nabla F_k(\mathbf{w}_k^\tau) $, thus $\mathbb{E}[\mathbf{g}_\tau]=\overline{\mathbf{g}}_\tau$.

\begin{table}[h!]
\centering
\caption{Notations and their descriptions.}
\begin{tabular}{|l|l|}
\hline
\textbf{Notation} & \textbf{Description} \\ \hline
$\mathbf{w}_k^\tau$ & Client $k$'s model weight at step $\tau$ \\ \hline
$\xi_k^\tau$ & Client $k$'s sampled data at step $\tau$ \\ \hline
$\eta_\tau$ & Learning rate at step $\tau$ \\ \hline
$m^{\tau+1}$ & Mean of global weights before shift, i.e., $\mathbb{E}[\mathbf{w}_t - \eta_t \overline{\mathbf{g}}_\tau]$ \\ \hline
$N$ & Total number of clients \\ \hline
$I$ & Number of inferior clients \\ \hline
$\alpha$ & $I/N$, the ratio of inferior clients \\ \hline
$p_k$ & Aggregation weight for client $k$ \\ \hline
$P$ &  The number of weight parameters in $\mathbf{w}$  \\ \hline
$\vec{\mathbf{1}}_P$ & $P$-sized vector with every element 1\\ \hline
\end{tabular}
\label{table:notations}
\end{table}

\subsection{Key Lemmas}
We first establish these \textbf{Lemma}s to be used in the proof of \textbf{Theorem} \ref{th1}.
\begin{lemma}\label{lemma1}
    Assuming that Assumption \ref{AS3} holds and if $\eta_\tau \leq \frac{1}{4L}$, following inequality holds.
   \begin{align}
    &\mathbb{E}[\|\overline{\mathbf{w}}^{\tau+1} - \mathbf{w}^*\|^2] \nonumber\\ 
    &\leq \left(1 + \frac{1}{\varepsilon}\right) \Big( (1 - \eta_\tau \mu) \mathbb{E}[\|\overline{\mathbf{w}}^\tau - \mathbf{w}^*\|^2] 
        + 6L\eta_\tau^2 \Gamma \nonumber \\ 
    &\quad + 2\mathbb{E}\bigg[ \sum_{k=1}^N p_k \|\overline{\mathbf{w}}^\tau - \mathbf{w}_k^*\|^2 \bigg] \Big) 
        + \eta_\tau^2 \mathbb{E}[\|\mathbf{g}_\tau - \overline{\mathbf{g}}_\tau\|^2] \nonumber \\
    &\quad + \left(1 + \varepsilon\right) \|\vec{\mathbf{1}}_P\|^2 \mathbb{E}[(\alpha m^{t+1})^2] \nonumber
\end{align}

\end{lemma}

We recall \textit{Lemma 2} and \textit{3} of \cite{li2019convergence} as follows.
\begin{lemma}\label{lemma2}
    Assuming that Assumption \ref{AS3} holds, it follows that
    \begin{equation}
    \mathbb{E}[\left\|\mathbf{g}_\tau-\overline{\mathbf{g}}_\tau\right\|^2 ]\leq \sum_{k=1}^N p_k^2 \sigma_k^2 .   
    \end{equation}
\end{lemma}

\begin{lemma}\label{lemma3}
    Assuming that Assumption \ref{AS4} holds and $\eta_\tau$ is non-increasing and $\eta_\tau\leq 2\eta_{\tau+E}$ for all $t$, the following inequality satifies.
    \begin{equation}
    \mathbb{E}\left[\sum_{k=1}^N p_k\left\|\overline{\mathbf{w}}^\tau-\mathbf{w}_k^\tau\right\|^2\right] \leq 4 \eta_\tau^2(E-1)^2 G^2   
    \end{equation}
\end{lemma}

\subsection{Proof of Lemma 1}
\begin{proof} We have global update rule for \texttt{FedShift} as $ \overline{\mathbf{w}}^{\tau+1} = \overline{\mathbf{w}}^\tau - \eta_\tau \mathbf{g}_\tau - \alpha m^{\tau+1} \vec{\mathbf{1}}_P $, therefore,

\begin{align}
&\| \overline{\mathbf{w}}^{\tau+1} - \mathbf{w}^* \|^2 = \| \overline{\mathbf{w}}^\tau - \eta_\tau \mathbf{g}_\tau - \alpha m^{\tau+1} \vec{\mathbf{1}}_P - \mathbf{w}^* \|^2 \nonumber\\ 
&= \| \overline{\mathbf{w}}^\tau - \mathbf{w}^* - \eta_\tau \mathbf{g}_\tau - \alpha m^{\tau+1} \vec{\mathbf{1}}_P - \eta_\tau \overline{\mathbf{g}}_\tau + \eta_\tau \overline{\mathbf{g}}_\tau \|^2 \nonumber\\ 
&= \underbrace{\| \overline{\mathbf{w}}^\tau - \mathbf{w}^* - \eta_\tau \overline{\mathbf{g}}^\tau - \alpha m^{\tau+1} \vec{\mathbf{1}}_P \|^2}_{Q_1} + \eta_\tau^2 \| \mathbf{g}_\tau - \overline{\mathbf{g}}_\tau \|^2 \nonumber\\ 
&\quad + \underbrace{2\eta_\tau \langle \overline{\mathbf{w}}^\tau - \mathbf{w}^* - \eta_\tau \overline{\mathbf{g}}_\tau - \alpha m^{\tau+1} \vec{\mathbf{1}}_P, \mathbf{g}_\tau - \overline{\mathbf{g}}_\tau \rangle }_{Q_2} \nonumber
\end{align}

According to \textbf{Lemma} \ref{lemma2}, the second term on RHS is bounded as $\eta_\tau^2 \| \mathbf{g}_\tau - \overline{\mathbf{g}}_\tau \|^2 \leq \eta_\tau^2\sum_{k=1}^N p_k^2 \sigma_k^2$.
Note that when we take expectation, $\mathbb{E}[Q_2]=0$ due to unbiased gradient.

Then, We expand $Q_1$

\begin{align}
&Q_1 = \| \overline{\mathbf{w}}^\tau - \mathbf{w}^* - \eta_\tau \overline{\mathbf{g}}_\tau - \alpha m^{\tau+1} \vec{\mathbf{1}}_P \|^2 \nonumber \\
&= \underbrace{\| \overline{\mathbf{w}}^\tau - \mathbf{w}^* - \eta_\tau \overline{\mathbf{g}}_\tau \|^2}_{A_1} + (\alpha m^{\tau+1})^2 \| \vec{\mathbf{1}}_P \|^2 \nonumber \\
& - 2 \langle \overline{\mathbf{w}}^\tau - \mathbf{w}^* - \eta_\tau \overline{\mathbf{g}}_\tau, \alpha m^{\tau+1} \vec{\mathbf{1}}_P \rangle \nonumber \\
&= A_1 + (\alpha m^{\tau+1})^2 \| \vec{\mathbf{1}}_P \|^2 - 2 \langle \overline{\mathbf{w}}^\tau - \mathbf{w}^* - \eta_\tau \overline{\mathbf{g}}_\tau, \alpha m^{\tau+1} \vec{\mathbf{1}}_P \rangle \nonumber \\ \label{IE1}
&\leq A_1 + \frac{1}{\varepsilon} A_1 + \varepsilon (\alpha m^{\tau+1})^2 \|\vec{\mathbf{1}}_P\|^2 + (\alpha m^{\tau+1})^2 \|\vec{\mathbf{1}}_P\|^2 \nonumber \\ 
&= \left(1 + \frac{1}{\varepsilon}\right) A_1 + \left(1 + \varepsilon\right) (\alpha m^{\tau+1})^2 \|\vec{\mathbf{1}}_P\|^2 \nonumber 
\end{align}

where inequality (\ref{IE1}) is due to a useful inequality property as follows.
\begin{align}
&\text{(by AM-GM, Cauchy-Schwartz, } \epsilon > 0): \\
&-2\langle a, b \rangle \leq \epsilon \|a\|^2 + \epsilon^{-1} \|b\|^2 \quad 
\end{align}
which make the cross term on RHS satisfy the inequality below
\begin{align*}
& - 2 \langle \overline{\mathbf{w}}^\tau - \mathbf{w}^* - \eta_\tau \overline{\mathbf{g}}_\tau, \alpha m^{\tau+1} \vec{\mathbf{1}}_P \rangle \\
&\leq \frac{1}{\varepsilon} \| \overline{\mathbf{w}}^\tau - \mathbf{w}^* - \eta_\tau \overline{\mathbf{g}}_\tau \|^2 + \varepsilon (\alpha m^{\tau+1})^2 \|\vec{\mathbf{1}}_P\|^2.
\end{align*}

Then, $A_1$ is known to be bounded as follows (see \cite{li2019convergence} for detailed proof).

\begin{align*}
A_1 &\leq \left(1-\mu \eta_\tau\right) \| \overline{\mathbf{w}}^\tau - \mathbf{w}^*\|^2 + 2 \sum_{k=1}^N \|\overline{\mathbf{w}}^\tau - \mathbf{w}_k^*\|^2   + 6\eta_\tau^2 L \Gamma
\end{align*}

Therefore, writing above inequalities and \textbf{Lemma} \ref{lemma2} and \ref{lemma3} all together, we have

\begin{align}
    &\mathbb{E}[\|\overline{\mathbf{w}}_{t+1} - \mathbf{w}^*\|^2] \leq 
    \left(1 + \frac{1}{\varepsilon}\right) \bigg((1 - \eta_\tau \mu) 
    \mathbb{E}[\|\overline{\mathbf{w}}^\tau - \mathbf{w}^*\|^2] \nonumber \\
    &+ 6L\eta_\tau^2 \Gamma + 2\mathbb{E}\left[ \sum_{k=1}^N p_k 
    \|\overline{\mathbf{w}}^\tau - \mathbf{w}_k^*\|^2 \right]\bigg) \nonumber \\
    &+ \eta_\tau^2 \mathbb{E}[\|\mathbf{g}_\tau - \overline{\mathbf{g}}_\tau\|^2] 
    + \left(1 + \varepsilon\right) \|\vec{\mathbf{1}}_P\|^2 
    \mathbb{E}[(\alpha m_{t+1})^2] \nonumber
\end{align}

\end{proof}

\subsection{Proof of Theorem \ref{th1}}
\begin{proof}

Let \( \Delta_\tau = \mathbb{E}\|\overline{\mathbf{w}}^\tau - \mathbf{w}^*\|^2 \) and \( \|\vec{\mathbf{1}}_P\| = P \)  and we begin with the result of \textbf{Lemma} \ref{lemma1}.

\begin{align}
&\Delta_{\tau+1} \leq (1 + \frac{1}{\varepsilon}) (1 - \eta_\tau \mu) \Delta_\tau + B_1\eta_\tau^2  + B_2 \mathbb{E}[(m^{\tau+1})^2] \nonumber\\ 
&\text{where } B_1 = (1 + \frac{1}{\varepsilon}) 6L\Gamma + (1 + \frac{1}{\varepsilon}) 8(E-1)^2 G^2 + \sum_{k=1}^N p_k \sigma_k^2 \nonumber\\
&\text{and }  B_2 = (1 + \varepsilon) P\alpha^2 \nonumber
\end{align}

To prove the convergence of the solution, we aim to make LHS be bounded by the function of total round $\tau=ET$ by induction.
For a diminishing step size, $\eta_\tau = \frac{\beta}{\tau+\gamma}$ for some $\beta >\frac{1}{\mu}$ and $\gamma>0$ such that $\eta_1 \leq \min\{\frac{1}{\mu}, \frac{1}{4L}\}=\frac{1}{4L}$ and $\eta_\tau \leq 2\eta_{\tau+E}$.
We will prove $\Delta_{\tau}\leq \frac{\upsilon}{\tau+\gamma}$ for all $\tau$ where $\displaystyle \upsilon =\max \left\{-\frac{B_1\beta^2 +B_2 M (\tau+\gamma)^2}{1-(1 + \frac{1}{\varepsilon})\beta\mu+\frac{1}{\varepsilon}(\tau+\gamma)},(\gamma+1)\Delta_1 \right\}$.

First, when $\tau = 1$, it holds due to the definition of $\upsilon$. Then, assuming $\Delta_{\tau}\leq \frac{\upsilon}{\tau+\gamma}$ holds, it follows that

\begin{align}
&\Delta_{\tau+1} \leq \left(1 + \frac{1}{\varepsilon}\right)(1 - \eta_\tau \mu)\Delta_\tau + B_1\eta_\tau^2  + B_2 \mathbb{E}[ (m^{\tau+1})^2] \nonumber \\
&\stackrel{(1)}{\leq} \left(1 + \frac{1}{\varepsilon}\right)(1 - \eta_\tau \mu)\Delta_\tau + B_1\eta_\tau^2 + B_2 M \nonumber \\
&\stackrel{(2)}{\leq} \frac{\left(1 + \frac{1}{\varepsilon}\right)(t+\gamma-\beta\mu)\upsilon 
+ B_1\beta^2  + B_2 M(t+\gamma)^2}{(t+\gamma)^2} \nonumber \\
&= \frac{(t+\gamma - 1)\upsilon + (1 - \beta \mu)\upsilon 
+ \frac{1}{\varepsilon}(t+\gamma - \beta \mu)\upsilon}{(t+\gamma)^2} \nonumber \\
&\quad + \frac{B_1\beta^2  + B_2 M(t+\gamma)^2}{(t+\gamma)^2} \nonumber \\
&= \frac{t+\gamma - 1}{(t+\gamma)^2}\upsilon \nonumber \\
&\quad + \frac{\left(1 - (1+\frac{1}{\varepsilon})\beta \mu 
+ \frac{1}{\varepsilon}(t+\gamma)\right)\upsilon + B_1 \beta^2 
+ B_2 M(t+\gamma)^2}{(t+\gamma)^2} \nonumber \\
&\stackrel{(3)}{\leq} \frac{\upsilon}{\tau+\gamma+1}.
\end{align}

where inequality (1) follows from \textbf{Assumption} \ref{AS5}, and inequality (2) follows because we have $\eta_\tau = \frac{\beta}{\tau+\gamma}$ and assumed $\Delta_{\tau}\leq \frac{\upsilon}{\tau+\gamma}$.
Also, inequality (3) follows as $\frac{\tau+\gamma-1}{(\tau+\gamma)(\tau+\gamma)} < \frac{\tau+\gamma-1}{(\tau+\gamma+1)(\tau+\gamma-1)}$ holds. Therefore, $\Delta_{\tau}\leq \frac{\upsilon}{\tau+\gamma}$ holds for all $\tau=1, 2, \cdots$.

Then, by $L$-smooth of $F$

\begin{align}
&\mathbb{E}[F(\overline{\mathbf{w}}^\tau)] - F^* \nonumber \\
&\leq \frac{L}{2} \Delta_\tau \leq \frac{L}{2} \frac{\upsilon}{\tau+\gamma} \nonumber \\
&\stackrel{(4)}{\leq} \frac{L}{2} \frac{1}{\tau+\gamma} 
\Bigg( \frac{B_1\beta^2 + B_2 M(\tau+\gamma)^2}{\left(1 + \frac{1}{\varepsilon}\right)\beta \mu 
- \frac{1}{\varepsilon}(\tau+\gamma)-1} 
+ (\gamma + 1) \Delta_1 \Bigg) \nonumber \\
&= \frac{L}{2} \Bigg[ \frac{1}{\tau+\gamma} 
\Bigg( \frac{B_1 \beta^2 }{\left(1 + \frac{1}{\varepsilon}\right)\beta \mu 
- \frac{1}{\varepsilon}(\tau+\gamma)-1} \nonumber \\
&\quad + \frac{B_2 M(\tau+\gamma)^2}{\left(1 + \frac{1}{\varepsilon}\right)\beta \mu 
- \frac{1}{\varepsilon}(\tau+\gamma)-1} \Bigg) 
+ \frac{1}{\tau+\gamma} (\gamma + 1) \Delta_1 \Bigg] \nonumber \\
&= \underbrace{\frac{L}{2} \frac{1}{\tau+\gamma} 
\Bigg( \frac{B_1 \beta^2 }{\left(1 + \frac{1}{\varepsilon}\right)\beta \mu 
- \frac{1}{\varepsilon}(\tau+\gamma)-1} + (\gamma + 1) \Delta_1 \Bigg)}_{\text{Vanishing term}} \nonumber \\
&\quad + \underbrace{\frac{L}{2} \frac{B_2 M}{\left(1 + \frac{1}{\varepsilon}\right)\beta \mu 
- \frac{1}{\varepsilon} - \frac{1}{\tau+\gamma}}}_{\text{Vanishes when $M \to 0$}}.
\end{align}

where inequality (4) follows because we have 
\begin{align}
\displaystyle \upsilon =\max \left\{-\frac{B_1\beta^2 +B_2 M (\tau+\gamma)^2}{1-(1 + \frac{1}{\varepsilon})\beta\mu+\frac{1}{\varepsilon}(\tau+\gamma)},(\gamma+1)\Delta_1 \right\} \nonumber \\
\leq \frac{B_1\beta^2 + B_2 M(\tau+\gamma)^2}{\left(1 + \frac{1}{\varepsilon}\right)\beta \mu - \frac{1}{\varepsilon} \left(\tau+\gamma\right)-1} + (\gamma + 1) \Delta_1 \nonumber 
\end{align}

With the choice of $\beta = \frac{2}{\mu}$, $\gamma = \max\{ \frac{8L}{\mu}, E\}-1$, then $\eta_\tau=\frac{2}{\mu}\frac{1}{\gamma+\tau}$ we can verify that conditions for the \textbf{Lemma}s are satisfied, i.e., $\eta_\tau \leq \frac{1}{4L}$ and $\eta_\tau \leq 2\eta_{\tau+E}$. As $\tau$ increases, the first term vanishes and the second term converges to $\displaystyle -\frac{\varepsilon L B_2 M}{2}$ but it also vanishes when $M \rightarrow 0$ which is the most abundant case in practice.

\end{proof}

\section{Weights Divergence Analysis}\label{DAproof}

Here, our goal is to measure how much \textit{the updated global model} diverges from \textit{the previous round's global model} to examine the effect of weight shifting. 
L2 distance of \texttt{FedAvg} divergence and \texttt{FedShift} divergence are examined for comparison.
Let $\Delta \mathbf{w}_k^t$ be the net difference between the distributed model and the model after local updates as follows.
\begin{equation}
    \Delta \mathbf{w}_k^t=\mathbf{w}^{t+1}_k-\mathbf{w}^{t}
\end{equation}
The model weights $\mathbf{w}^{t+1}_k$ include any changes that is made by local updates, for example, aggregations weights would be included in this term.
Here, we analyze the weights within a single round, therefore, we drop superscript $t$ index, i.e., $\Delta \mathbf{w}_k$.
Then, the weight aggregation of any FL algorithms would be generally expressed as

\begin{equation}
    \displaystyle \mathbf{w}^{t+1} = \mathbf{w}^{t}+\frac{1}{K}\sum_k \Delta \mathbf{w}_k , \\
\end{equation}

whereas our \texttt{FedShift} algorithm would be expressed as 

\begin{align}
    \displaystyle \mathbf{\hat{w}}_{t+1} &= \mathbf{w}^{t+1}-\frac{I}{K}m^{t+1}  \vec{\mathbf{1}}_P \\
    \displaystyle &= \mathbf{w}^{t}+\frac{1}{K}\sum_K \Delta w_k-\frac{I}{K}m^{t+1}  \vec{\mathbf{1}}_P.
\end{align}

\subsection{Proof of Theorem \ref{th2}}
\begin{proof}

Now, we compare L2 distance of
$(\mathbf{w}_g^{t+1} : \mathbf{w}_g^{t})$ and $(\mathbf{\hat{w}}_{g}^{t+1} : \mathbf{w}_g^{t})$ which are denoted as $D_{FA}$ and $D_{FS}$, respectively.
\begin{equation}
\begin{array}{l}
    \displaystyle D_{FA}^2=||\mathbf{w}^{t+1} - \mathbf{w}^{t}||_2^2 =(\mathbf{w}^{t+1} - \mathbf{w}^{t})^\intercal(\mathbf{w}^{t+1} - \mathbf{w}^{t})\\
    \displaystyle = \left(\frac{1}{K}\sum_k \Delta \mathbf{w}_k \right)^2
\end{array}
\end{equation}

\begin{equation}
\begin{array}{l}
    \displaystyle D_{FS}^2=||\mathbf{\hat{w}}^{t+1} - \mathbf{w}^{t}||_2^2 
    =(\mathbf{\hat{w}}_g^{t+1} - \mathbf{w}^{t})^\intercal(\mathbf{\hat{w}}^{t+1} - \mathbf{w}^{t})\\
    \displaystyle  = \left(\frac{1}{K}\sum_k \Delta \mathbf{w}_k-\frac{I}{K}m^{t+1}\Vec{\mathbf{1}}_P\right)^2 \\
    \displaystyle  =  \left(\frac{1}{K}\sum_k \Delta \mathbf{w}_k \right)^2 + \left(\frac{I}{K}m^{t+1} \Vec{\mathbf{1}}_P\right)^2 \\
    - 2\frac{Im^{t+1}}{K^2}\sum_k \Delta \mathbf{w}_k \cdot \Vec{\mathbf{1}}_P
\end{array}
\end{equation}

To compare the value of $D_{FA}$ and $D_{FS}$, we subtract them as follows

\begin{equation}\label{DA1}
\begin{array}{l}
    \displaystyle D_{FA}^2-D_{FS}^2 = 2\frac{Im^{t+1}}{K^2}\sum_k \Delta \mathbf{w}_k \cdot \Vec{\mathbf{1}}_P - \left(\frac{I}{K}m^{t+1} \Vec{\mathbf{1}}_P\right)^2
\end{array}
\end{equation}

From the definition of $m^{t+1}$ we can also represent it as a dot product of vectors as follows.
\begin{equation}
\begin{array}{l}
    \displaystyle m^{t+1}=\frac{1}{P}\sum_p \mathbf{w}^{t+1}[p]=\frac{1}{P} \mathbf{w}^{t+1}\cdot \Vec{\mathbf{1}}_P\\
\end{array}
\end{equation}

Then, we have $\frac{1}{K}\sum_k \Delta \mathbf{w}_k \cdot \Vec{\mathbf{1}}_P = P(m^{t+1}-m^t)$ and $\Vec{\mathbf{1}}_P\cdot \Vec{\mathbf{1}}_P = P$ which are to be plugged into Eq. (\ref{DA1}).

\end{proof}

\begin{align}
    \displaystyle D_{FA}^2-D_{FS}^2 &= \frac{2IP}{K}m^{t+1}(m^{t+1}-m^t)- P\left(\frac{I}{K}m^{t+1} \right)^2\\
    &= \frac{IP}{K}\left((m^{t+1})^2(2-\frac{I}{K}) -2m^{t+1}m^{t}\right)
\end{align}

\textbf{1. Condition that \texttt{FedShift} divergence is smaller:} $D_{FA}-D_{FS}>0$

\begin{equation}
\begin{array}{l}
    \displaystyle \frac{K+S}{2K} > \frac{m^{t}}{m^{t+1}}\\
\end{array}
\end{equation}

\textbf{2. Condition that \texttt{FedShift} divergence is greater:} $D_{FA}-D_{FS}<0$

\begin{equation}
\begin{array}{l}
    \displaystyle \frac{K+S}{2K} < \frac{m^{t}}{m^{t+1}}\\
\end{array}
\end{equation}

To understand the above condition, let's take a simple example.
Assume $m^{t+1}>0$ and inferior and superior clients are 50\%  and 50\%; then, when $m^{t+1} > \frac{4}{3}m^{t}$ holds, the divergence of Fedshift's update is smaller than that of \texttt{FedAvg}'s update.
When $m^{t+1} < \frac{4}{3}m^{t}$ holds, the divergence of Fedshift's update is greater than that of \texttt{FedAvg}'s update.
If the average of weights is updated to go beyond the threshold, e.g., $\frac{4}{3}$ of the previous average, it decreases the distance from the previous model. If the average of weights is updated to be within the threshold, e.g., $\frac{4}{3}$ of the previous average, it increases the distance from the previous model.
It can be seen as letting the model be brave within the fence an rather be cautious outside of the fence.



%

\bibliography{main}
\bibliographystyle{IEEEtran}

\newpage

 




\vfill

\end{document}